\documentclass[12pt,a4paper]{report}
\usepackage{amsmath,amsthm,amssymb,graphicx,hyperref}
\usepackage[left=1.2in,right=1in,top=1in,bottom=1in]{geometry}
\usepackage[romanian]{babel}

\usepackage{amsmath}
\usepackage{graphicx}
\usepackage{float}
\graphicspath{{images/} {diagrams/}}

\usepackage{fullpage}

\usepackage{url}
\usepackage{listings}
\usepackage{verbatim}
\usepackage{color}
\usepackage[acronym]{glossaries}
\usepackage{subcaption}

\usepackage{biblatex}
\usepackage{multirow}
\usepackage{array}
\theoremstyle{definition}

\theoremstyle{remark}

\begin{document}
\thispagestyle{empty}
\begin{center}
\begin{figure}[h!]
\vspace{-20pt}
\begin{center}
\includegraphics[width=100pt]{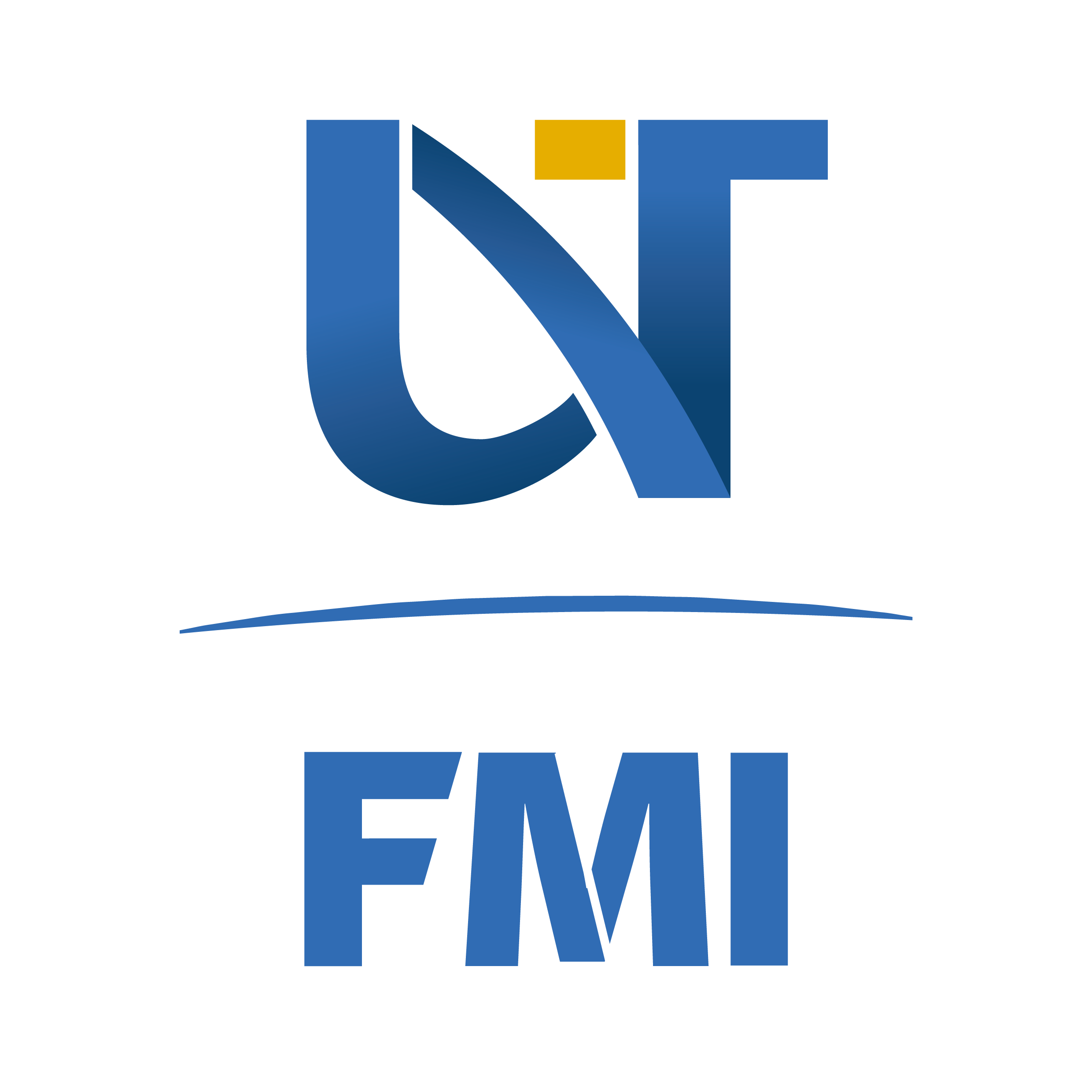}
\end{center}
\end{figure}

{\large{\bf UNIVERSITATEA DE VEST DIN TIMI\c SOARA

FACULTATEA DE MATEMATIC\u A \c SI INFORMATIC\u A

PROGRAMUL DE STUDII DE LICEN\c T\u A: Informatic\u a }}

\vspace{120pt}
{\huge {\bf LUCRARE DE LICEN\c T\u A}}

\vspace{150pt}
\end{center}

{\large\noindent{\bf COORDONATOR:\hfill ABSOLVENT:}

\noindent Conf.Dr. Darian  \textsc{Onchiș} \hfill Hogea Eduard Florin}

\vfill
\begin{center}
{\bf TIMI\c SOARA

2022}
\end{center}
\newpage
\thispagestyle{empty}
\begin{center}
{\large{\bf UNIVERSITATEA DE VEST DIN TIMI\c SOARA

FACULTATEA DE MATEMATIC\u A \c SI INFORMATIC\u A

PROGRAMUL DE STUDII DE LICEN\c T\u A : Informatic\u a }}

\vspace{120pt}
{\huge {\bf SISTEME HIBRIDE DE ÎNVĂȚARE AUTOMATĂ ȘI APLICAȚII}} 

\vspace{150pt}
\end{center}

{\large\noindent{\bf COORDONATOR:\hfill ABSOLVENT:}

\noindent Conf.Dr. Darian  \textsc{Onchiș} \hfill Hogea Eduard Florin}

\vfill
\begin{center}
{\bf TIMI\c SOARA

2022}
\end{center}
\newpage
\normalsize{}
\section*{Abstract}

In this paper, a deep neural network approach and a neuro-symbolic one are proposed for classification and regression. The neuro-symbolic predictive models based on Logic Tensor Networks are capable of discriminating and in the same time of explaining the characterization of bad connections, called alerts or attacks, and of normal connections. The proposed hybrid systems incorporate both the ability of deep neural networks to improve on their own through experience and the interpretability of the results provided by symbolic artificial intelligence approach. To justify the need for shifting towards hybrid systems, explanation, implementation, and comparison of the dense neural network and the neuro-symbolic network is performed in detail. For the comparison to be relevant, the same datasets were used in training and the metrics resulted have been compared. A review of the resulted metrics shows that while both methods have similar precision in their predictive models, with Logic Tensor Networks being also possible to have interactive accuracy and deductive reasoning over data. Other advantages and disadvantages such as overfitting mitigation and scalability issues are also further discussed.

\tableofcontents
\newpage

\chapter{Introducere} 

\pagenumbering{arabic}
\setcounter{page}{1}

\section{Motivație și scop} 
În prezent, inteligența artificială reprezintă cel mai dezbătut și popular domeniu al informaticii. Atenția asupra acestui subiect poate fi justificată prin potențialul și multitudinea de domenii de aplicabilitate. Termenul de inteligență artificială nu este însă așa de recent precum s-ar putea crede, originile acestuia datând de la mijlocul secolului trecut. Cu toate acestea, interesul pentru inteligența artificială nu a fost unul constant. Puterea de procesare necesară, insuficiența volumului de date și eticitatea au fost, și încă sunt factori limitatori. Vasta sa aplicabilitate și atenția oferită au făcut posibilă folosirea inteligenței artificiale chiar și în combaterea pandemiei actuale. Un exemplu concret îl constituie monitorizarea și abilitatea de a prezice cu o acuratețe semnificativă zonele susceptibile de a deveni focare de infecție. O ramură a inteligenței artificiale care face aceste lucruri posibile o constituie învățarea automată (machine learning) și abordările recente ale acesteia sub forma Învățării Profunde(Deep Learning). Scopul acestei lucrări este de a justifica importanța unui algoritm care conține atât logica oferită de Symbolic AI și potențialul de dezvoltare al antrenării, comparând rezultatele cu cele obținute prin implementarea algoritmilor de Învățare Profundă. 

\section{Contextul lucrării}

În această lucrare este realizat un studiu comparativ al celor două abordări și sunt prezentate avantajele și dezavantajele acestora. Au fost realizate  modele predictive de regresie și de clasificare cu ajutorul unui algoritm neuro-symbolic bazat pe logica de oridinul întâi și a unuia bazat pe o rețea neuronală profundă. Ca algoritm neuro-symbolic a fost ales Logic Tensor Network, ținându-se cont de potențialul impact, relevanța în domeniu și experiența proprie în aplicarea acestuia iar pentru cel de învățare profundă a fost folosită o rețea neuronală profundă.
Un alt aspect luat în considerare care stă la baza comparației îl constituie selecția algoritmilor de dificultate diferită. În acest caz, sistemul hibrid necesită o atenție deosebită, fiind necesară pe lângă crearea unei rețele și antrena-re a acesteia și adăugarea unei părți de raționament logic.
Pentru o mai bună înțelegere, este oferită o descriere detaliată a modului de funcționare atât prin explicații cât și prin diverse diagrame UML. Pe lângă această parte teoretică, lucrarea conține și o parte de experimente numerice, realizată cu scopul de a vedea diferențele dintre cele două moduri de rezolvare a problemelor de clasificare și regresie. În antrenarea acestor algoritmi, s-au păstrat măsurători clare asupra preciziei, satisfiabilității, al numărului de epoci folosite precum și al altor aspecte relevante într-o lucrare de sinteză.

În metodologia de realizare explicată ulterior sunt descrise modul de implementare, bibliotecile utilizate împreună cu specificarea versiunii lor și mediile de dezvoltare din realizarea experimentelor numerice. Alături de acestea este și o prezentare a instrumentelor software folosite, însoțite de un ghid despre posibila reproducere a rezultatelor. 

\section{Soluții similare}

Necesitatea automatizării în anumite domenii și recunoașterea potențialului algoritmilor avansați de învățare automată au dus la apariția multor lucrări. O lucrare similară prin modul analitic al algoritmilor o constituie  \cite{palvanov2018comparisons}. Asemănările în scopul lucrării și modul de realizare între lucrarea citată și aceasta sunt numeroase, însă o diferență importantă o constituie tipul datelor folosite în setul de date. În articolul citat se folosește în antrenare și clasificare setul de date MNIST \cite{mnist}. 

MNIST cuprinde într-o cantitate foarte mare, imagini de dimensiuni mici cu numere de la 0 la 9 scrise de mână. Popularitatea acestui set de date este dată de simplitatea folosirii, fără a fi nevoie de o preprocesare considerabilă. Algoritmii folosiți sunt, precum în această lucrare, algoritmi avansați de învățare automată. Sunt realizate comparații între rețele neuronale care folosesc regresie logistică, rețele neuronale convoluționale, rețele reziduale și rețele neuronale cu capsule. Aspecte evidențiate în lucrarea menționată sunt memoria utilizată, acuratețea și timpul necesar pentu antrenare și evaluare.

Cu toate acestea, în această lucrare unul dintre seturile de date folosite este KDD99 \cite{kdd99}. Un set de date de dimensiuni mari care conține informații despre traficul dintr-o rețea. Clasificările realizate de către algoritmii analizați în această lucrare sunt pentru a distinge tipul conexiunii. Mai specific, pentru a distinge dacă o conexiune este normală sau dacă este un atac (precizând și tipul acestuia). Alte seturi de date folosite în studiul comparativ sunt CIC-IDS2017 și cel de detecție a deteriorării prin vibrații a une grinzi, furnizat de către Universitatea Babes-Bolyai din Cluj-Napoca.

O altă lucrare similară pe care o consider relevantă este \cite{nikou2019stock}. În esență fiind tot o lucrare de sinteză care detaliază diferențele dintre algoritmi cu scopul prezicerii prețurilor stocurilor. O asemănare importantă de notat este comparația dintre algoritmi cu dificultăți de implementare diferite.

\section{Contribuție proprie}
Probabil cea mai importantă parte a unei lucrări o constituie cea în care autorul își prezintă contribuția. În cazul meu, contribuția constă în următoarele realizări:

\begin{enumerate}
    \item Studierea problemelor și deficitelor din inteligența artificială simbolică și din invățarea profundă.
    \item Justificarea abilităților rețelelor LTN de a avea raționemnt deductiv și acuratețe interactivă.
    \item Implementarea unei rețele LTN cu scopul recunoașterii atacurilor din setul de date KDD99 \cite{neuroadvantages}.
    \item Implementarea unei rețele neuronale profunde cu scopul recunoașterii atacurilor din setul de date KDD99.
    \item Implementarea unei rețele LTN cu scopul recunoașterii atacurilor din setul de date CIC-IDS2017 \cite{neurosecurity}.
    \item Compararea și explicarea seturilor de date și a bazelor de cunoștinte formate pentru acestea \cite{flavia}.
    \item Folosirea mai multor modele predictive precum clasificare multi-label, clasificare single-label și regresie \cite{fetril}.
    \item Determinarea avantajelor sistemelor hibride pe baza experimentelor efectuate.
    \item Studierea folosirii sistemelor hibride în scenarii reale din industrie \cite{canti}.
    \item Determinarea defectelor unei grinzi prin seturi de frecvențe naturale.
    \item Folosirea predicatelor din Real Logic pentru a implementa funcții de similaritate.
    \item Realizarea unui studiu comparativ al predicatelor folosite și rezultatele modelelor constrânse în antrenare de acestea.
    \item Testarea modelelor antrenate imitând scenarii reale din industrie, când volumul seturilor de date este mic prin folosirea k-fold cross validation.
    
\end{enumerate}

\newpage{}

\chapter{Descrierea problemei}
Succesul și interesul crescut în Inteligența Artificială și în Machine Learning nu pot fi contestate. Cu toate acestea, au apărut multe discuții despre potențialele probleme. O primă problemă întâlnită în tehnicile de învățare profundă \cite{lecun2015deep,schmidhuber2015deep}, precum rețelele neuronale profunde este că ele necesită atât putere de procesare considerabilă cât și un set foarte mare de date de antrenare pentru a oferi rezultate bune. O altă problemă adresată de cercetători \cite{szegedy2013intriguing} este legată de rezultatele eronate obținute în urmă unei mici alterări a imaginii pe care se face recunoașterea. Alterare insesizabilă pentru o persoană care privește imaginea, însă care face ca o recunoaștere să fie greșită. Această problema a adus critici asupra susceptibilității în cazul unor atacuri. Probabil însă cea mai persistentă problemă discutată este lipsa unei explicații despre cum acea rețea neuronală face unele conexiuni. Aceasta este o problemă cunoscută sub numele de "black box problem" în Inteligența Artificială.

Pentru a adresa mai bine această problemă, voi oferi un exemplu concret. Unul dintre primele programe pe care le face cineva care vrea să învețe să folosească rețele neuronale este o clasificare între câini și pisici. Țînând cont că pentru o rețea neuronală să fie rețea neuronală profundă, aceasta are nevoie de un strat de intrare, minim două straturi ascunse(cu toate că numărul de straturi ascunse pentru a considera o rețea profundă poate să difere în funcție de sursa de informare) și un strat de ieșire. Stratul de intrare cuprinde datele inițiale(în cazul nostru o poză cu un câine sau o pisică) care sunt oferite, iar stratul de ieșire cuprinde ceea ce vrem să aflăm(clasificarea - câine sau pisică). Problema ambiguității în modul de gândire al acestor rețele apare însă în nivelele ascunse. Acolo are loc și partea de învățare automată a rețelei, însă singurul rezultat clar pe care îl avem din această învățare este în ultimul strat. Fără a știi(sau cel puțin fără a înțelege) cum au fost făcute aceste decizii, problema "black box" a continuat să atragă atenția cercetătorilor.

O soluție adusă în rezolvarea acelor probleme prezente în învățarea profundă este de a implementa și o parte de raționament logic, precum cel întâlnit în Inteligența Artificială Simbolică \cite{haugeland1989artificial}. Printr-o reprezentare simbolică, ușor de înțeles, și cu reguli clar definite Inteligență Artificială Simbolică poate să reducă puterea de procesare și dimensiunea setului de date necesare. Îmbinarea celor două abordări propune eficientizarea antrenării, logica fiind descrisă încă de la început, contrar cu logica dobândită în timpul antrenării în cazul învățarii profunde. Problemele de ambiguitate și rubustete sunt de asemenea rezolvate.

Este important de evidențiat faptul că, prin crearea unor sisteme hibride, nu toate problemele existente sunt totuși rezolvate. Prin natura flexibilă a inteligenței artificiale, au apărut inevitabil și abordări cu scopuri morale greșite.  Riscurile cu privire la modurile malițioase în folosire continuă să fie un impediment în dezvoltare, însă acele aspecte nu vor fi abordate în această lucrare.

Implementările hibride și modul în care interacționează acestea au fost discutate, în timpul conferinței AAAI-2020 de către Henry Kautz \cite{linkyt}. La premierea acestuia, a ținut un discurs despre cum Inteligența Artificială a fost folosită în trecut în scopuri greșite, despre posibilele pericole pe care le prezintă și despre viitorul Inteligenței Artificiale. În cadrul ultimei părți menționate, cercetătorul american a făcut următoarea clasificare dspre cum interacționează rețelele neuronale cu sistemele simbolice:

\textbf{symbolic Neuro symbolic}. Datele inițiale sunt convertite în reprezentări vectoriale denumite "embeddings" prin modele predictive precum "doc2vec", "word2vec", "GloVe". Acestea sunt trecute prin mai multe straturi ascunse pentru ca în final ultimul strat să fie la rândul lui reprezentat sub formă simbolică.

\textbf{Symbolic[Neuro]}. AlphaGo și succesorul acestuia, AlphaZero sunt unele dintre exemplele unde se regăsește acest tip de interacțiune. AlphaZero a reușit performanțele ca în numai câteva ore de antrenare, să învingă cei mai buni jucători de Go. Ambele programe au la bază algoritmul de tip Monte Carlo, asistat în alegerea mutării următoare de către o metodă de învățare profundă.

\textbf{Neuro $\cup$ compile(Symbolic)}
Lucrarea \cite{lample2019deep} explică cum datele de intrare pot fi sub formă de logică matematică. Un exemplu clar ar fi cel în care datele de intrare reprezintă o integrală, iar ultimul strat al rețelei reprezintă calculul acesteia.

\textbf{Neuro $\longrightarrow$ Symbolic} \cite{mao2019neuro}. Propune o implementare sub forma unui model de învățare automată prin analiza unor imagini și a unor întrebări și răspunsuri. Primul strat al rețelei este compus din imagini, dar în același timp și din propoziții. Acestea sunt trecute printr-o rețea neuronală iar rezultatul este un raționament logic între obiectele din imagini.

\textbf{Neuro[Symbolic]}. Presupune o utilizare a raționamentului logic în interiorul rețelei neuronale. În timpul antrenării, informațiile din rețeaua neuronală sunt transformate într-o reprezentare simbolică împreună cu scopul, iar asupra acestora este utilizat raționa-mentul logic definit. Rezultatul este transmis înapoi rețelei neuronale pentru a continua antrenarea.

\textbf{Neuro$_{Symbolic}$}. Un mod de interacțiune omis inițial de către Henry Kautz însă adăugat ulterior. Această categorie cuprinde toate implementările care au logică de ordinul întâi încapsulată în tensori, în care rețele neuronale sunt folosite pentru a le interpreta. 

\section{Rețele neuronale profunde}

Învățarea profundă a fost inspirată de modul în care funcționează creierul uman. Neuronii din domeniul informatic au fost descriși inițial de către Walter Pitts și McCulloch că valori binare, o rețea din acești neuroni fiind capabilă să ajungă la o inferență logică. Ideea aceasta a fost revoluționară, dar în același timp, prin limitările impuse de valorile binare, a împiedicat dezvoltarea timp de aproape două decenii. Învățarea profundă a redevenit posibilă prin evoluția calculatoarelor și descoperirea tehnicii de propagare inversă(backpropagation). Pentru o bună vreme, a avut parte de mult atenție din partea cercetătorilor însă comercializarea Inteligenței Artificiale la acea vreme constă în Sistemele Expert. Dezvoltările învățării profunde în următorii ani au fost afectate, până în anii 2009-2010. Atunci au apărut implementări în recunoaștere vocală și recunoaștere a imaginilor prin rețele neuronale cu mai multe straturi ascunse care se foloseau de algoritmul de propagare inversă. Evoluția învățării profunde este accentuată de mai mulți factori. Puterea de procesare crescută, seturile de date de dimensiuni mari disponibile și îmbunătățirile algoritmilor existenți sunt doar o parte din acei factori.

Una dintre arhitecturile din învățarea profundă este cea de Rețele Neuronale Profunde. Figura 2.1 descrie o posibilă implementare a unei rețele neuronale profunde, cu două componente în straturile de intrare și de ieșire și câte 4 în cele ascunse.

\begin{figure}[!htbp]
    \centering
    \includegraphics[width=1\textwidth]{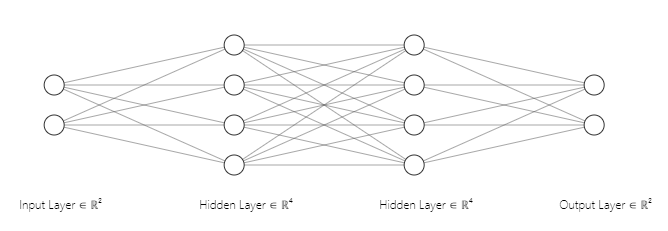}
    \caption{O rețea neuronală profundă simplă}
    \label{dnn}
\end{figure}

Rețelele Neuronale Profunde oferă soluții în foarte multe domenii, ele excelând prin abilitatea de a extrage caracteristici din setul de date și de a folosi învățarea automată pe ele. Implementări multiple ale acestor rețele ca metode de învățare supervizată, învățare nesupervizată, învățare supervizată parțial și învățare prin armare au fost descrise în diverse articole precum \cite{memon2020loop} \cite{jafari2021deep} .

Rezultatele foarte bune obținute de aceste rețele sunt de cele mai multe ori puternic influențate de numărul de straturi ascunse. Cu toate că acuratețea din antrenare crește și ea împreună adăugarea mai multor straturi ascunse, explicabilitate modului în care rețeaua a ajuns la acele rezultate scade. Odată cu ea, scade și abilitatea modelului rezultat din urmă antrenării de a generaliza, acesta devenind predispus la greșeli prin aplicarea pe un nou set de date.

Pentru a efectua un studiu comparativ relevant între această arhitectură și un sistem hibrid în care sunt definite și elementele logice, s-a ales să se folosească atât învățarea supervizată, întrucât în definirea bazei de cunoștinte și a predicatelor folosite s-au folosit anumite atribute din seturile de date.

\section{Logic Tensor Networks}
Logic Tensor Network(LTN) \cite{badreddine2022logic} este o abordare care scapă de deficitele din învățarea profundă și încorporează raționamentul logic. Țînând cont de clasificarea anterioară, s-ar încadra drept o implementare Neuro$_{Symbolica}$. Pentru a evita confuziile, în mențiunile și implementările ulterioare ale sistemelor hibride, se va folosi denumirea simplă de $neuro-symbolic$.

În LTN, integrarea logicii de ordinul întâi este realizată prin Real Logic. Acest lucru face posibilă aducerea în discuție a constantelor, variabilelor, obiectelor, funcțiilor, predicatelor, conectorilor, relațiilor și regulilor și este capabil de rezolvarea majorității problemelor din Inteligența Artificială. Autorii articolului original \cite{badreddine2022logic} au pus și la dispoziție exemple ale implementării clasificării binare, clasificării claselor mutiple cu o etichetă sau două, adunării cifrelor recunoscute din MNIST, a regresiei, clustering și un exemplu de predicție al legăturii dintre obiecte din setul de date Smokes Friends Cancer din lucrarea \cite{richardson2006markov}. 

Framework-ul neuro-symbolic se folosește de Real Logic prin transformarea acestuia în grafuri computaționale Tensorflow \cite{tensorflow2015-whitepaper}. \textbf{Constantele} folosite în Real Logic sunt sub forma unor tensori. Tensorii stau la baza majorității stucturilor de date folosite în învățarea automată. Aceștia pot avea orice rang, putând fi reprezentați ca o mulțime de orice dimensiune precum în urmatoarea formulă: 
$$ \bigcup_{i_1, i_2, ..., i_k \in \mathbb{N^*}} \mathbb{R}^{i_1 \times i_2 \times ... \times i_k} $$ 

Un tensor de rangul 0 este definit drept un scalar, rangul 1 fiind un vector iar de rangul 2 fiind o matrice. Exemplul poate să continuie.

\textbf{Predicatele} sunt funcții sau operații cu tensori care au ca rezultat o valoare cuprinsă între [0,1] interpretată ca nivel de satisfiabilitate. Această valoare de adevăr parțial poartă numele de logica fuzzy \cite{hajek2013metamathematics} și a fost folosită de mult timp în Inteligența Artificială Simbolică pentru a putea descrie situații vagi, unde răspunsul nu poate să fie unul cu certitudine fals sau adevărat. Variabile fuzzy precum "scurt", "rapid", "tânăr", "puternic" sunt exemple ale acestor concepte imprecise. \textbf{Conectorii} din Logic Tensor Network sunt folosiți pentru a lega predicatele definite anterior. Ei pot să fie negație, implicație, conjuncție, disjuncție iar rezultatul folosirii lor împreună cu predicatele este tot o valoare parțială de adevar. \textbf{Cuantificatorii} sunt de asemenea prezenți în această rețea neuro-symbolică, acestia fiind cuantificatorul existențial $\exists$ și cel universal $\forall$. \textbf{Variabilele} sunt definite ca secvențe de obiecte din aceeași mulțime. 

\begin{figure}[ht]
\centering
\includegraphics[width=14cm]{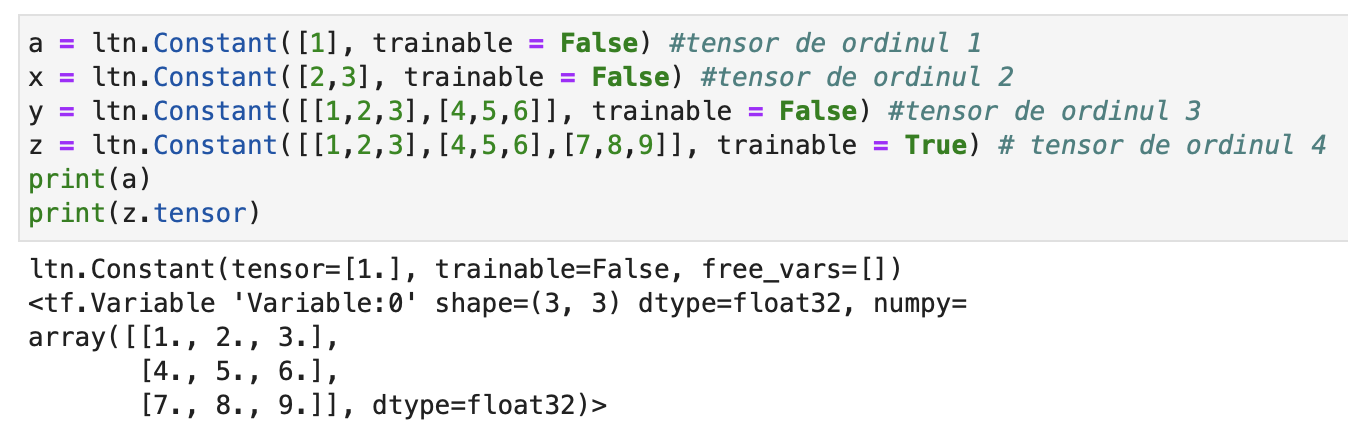}
\caption{Constante}
\label{const}
\end{figure}

În Figura 2.2 am arătat cum se inițializează tensori de ordin diferit și cum se afișează aceștia. Parametrul $trainable$ ne permite să definim caracteristicile ca fiind antrenabile. Dacă pentru o constantă $\alpha$ avem tensorul asociat $\mathbb{G}(\alpha)$, atunci în exemplul de cod din figura menționată avem următorii tensori $\mathbb{G}(a) ,\mathbb{G}(x), \mathbb{G}(y), \mathbb{G}(z)$ de ranguri diferite.

$ \mathbb{G}(a) = \begin{pmatrix}
1
\end{pmatrix}  $,
$ \mathbb{G}(x) = \begin{pmatrix}
2 & 3
\end{pmatrix}  $, $ \mathbb{G}(y) = \begin{pmatrix}
1 & 2 & 3\\
4 & 5 & 6
\end{pmatrix} $, $ \mathbb{G}(z) = \begin{pmatrix}
1 & 2 & 3\\
4 & 5 & 6\\
7 & 8 & 9\\
\end{pmatrix} $

\begin{figure}[ht]
\centering
\includegraphics[width=14cm]{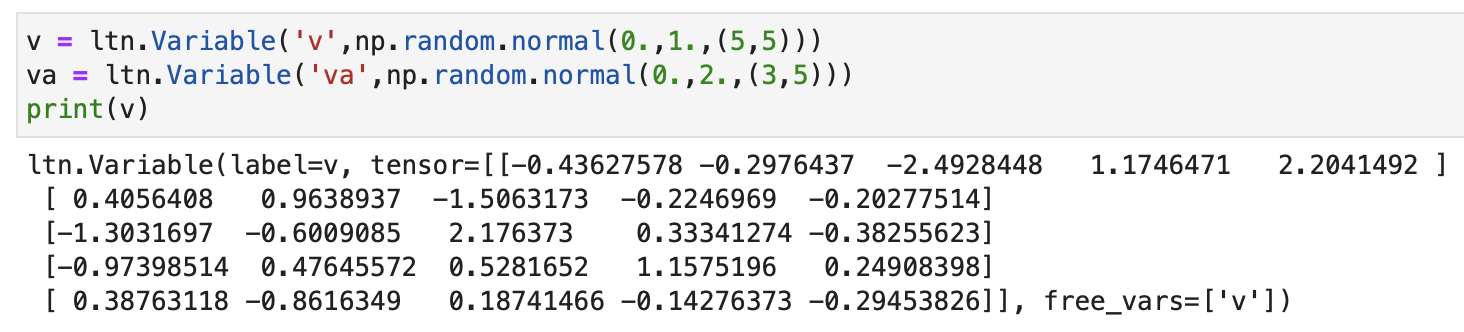}
\caption{Variabile}
\label{var}
\end{figure}

Un exemplu de inițializare a variabilelor din Logic Tensor Network este prezentat în Figura 2.3. Valoarea $v$ va avea în componență 5 numere din $R^2$ iar valoarea $va$ va avea 3. O afișare este de asemenea prezentată. Variabilele pot conține de asemenea constantele definite anterior.

\begin{figure}[ht]
\centering
\includegraphics[width=14cm]{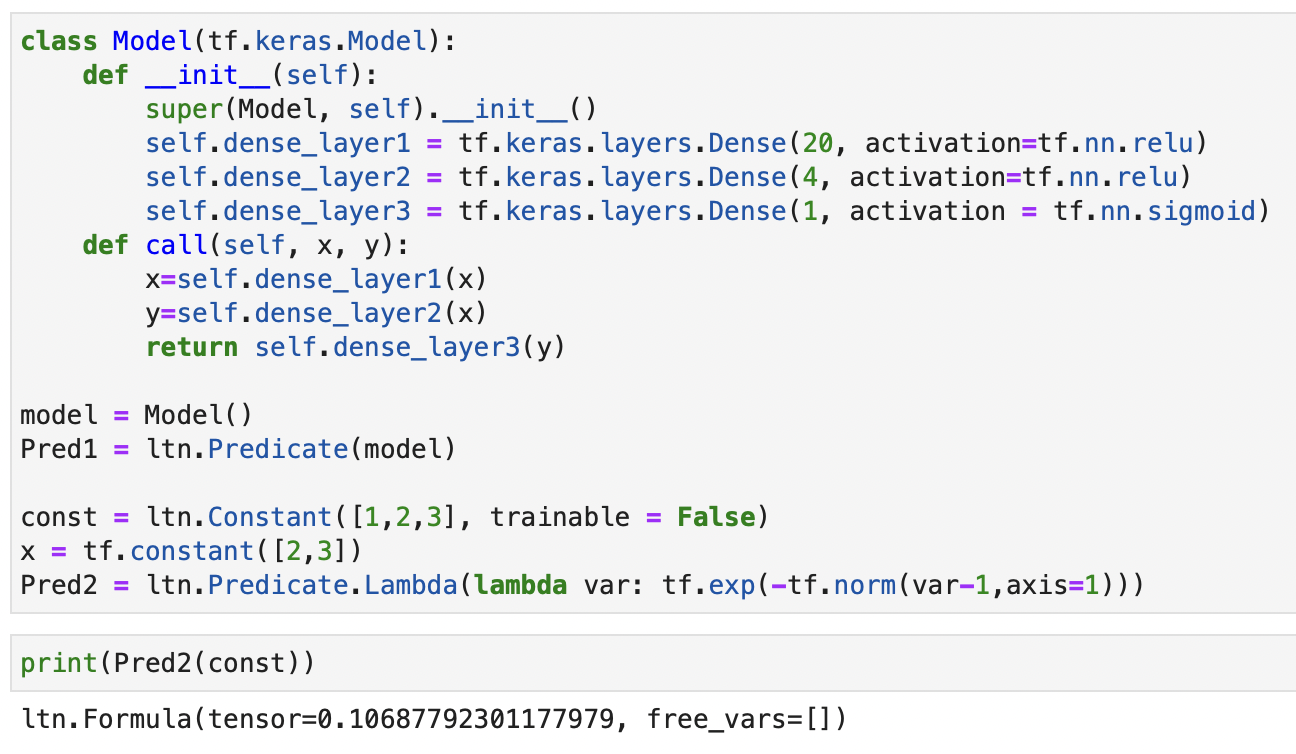}
\caption{Predicate}
\label{pred}
\end{figure}

Figura 2.4 arată o potențială folosire a predicatelor din LTN. Sunt prezentate un constructor care folosește implicit o rețea neuronală și o funcție $\lambda$. Predicatele definite sunt folosite pe constante iar rezultatul este o valoare de adevăr între valorile [0,1]. În cazul rețelei neuronale acest lucru este posibil prin adăugarea în ultimul strat a funcției de activare sigmoid.

\begin{figure}[ht]
\centering
\includegraphics[width=14cm]{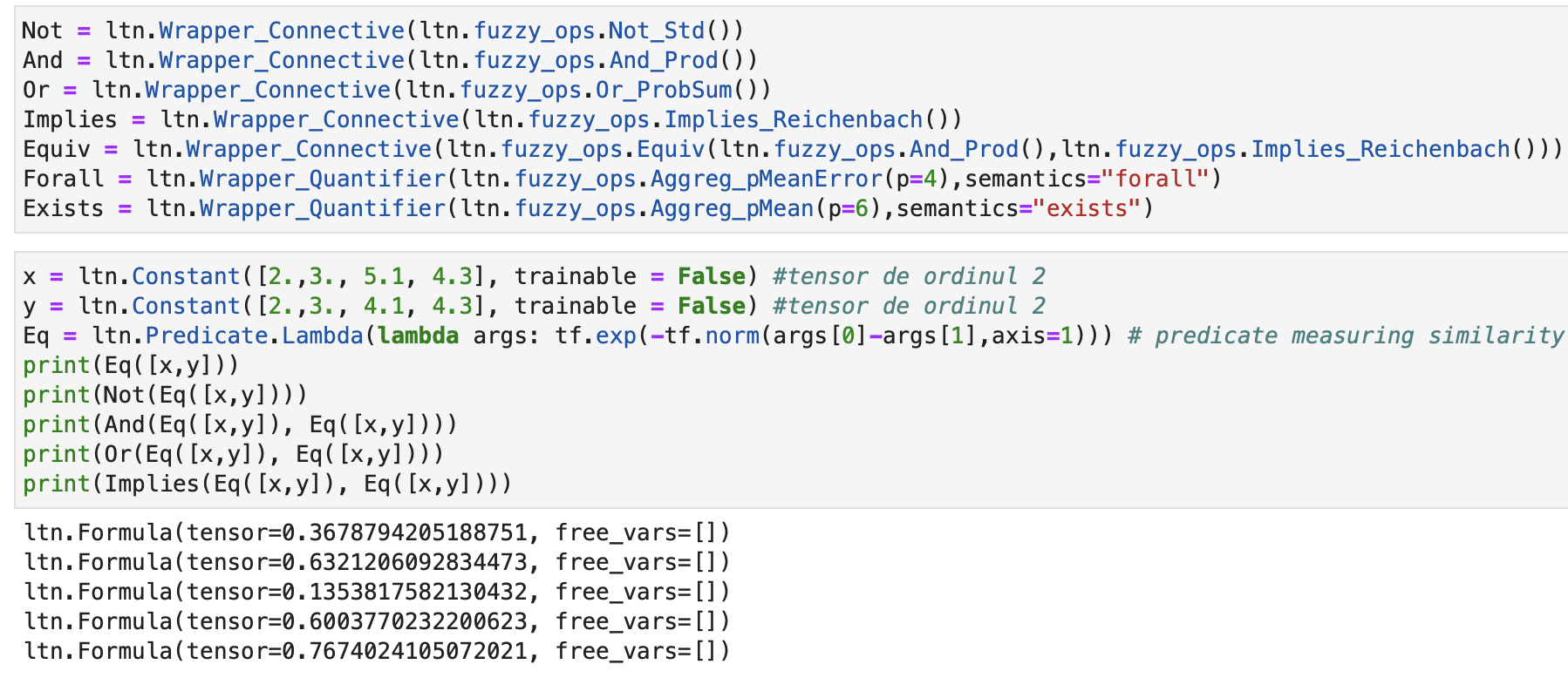}
\caption{Conectori și cuantificatori}
\label{con}
\end{figure}

Constantele, predicatele, conectorii și cuantificatorii sunt folosiți împreună în Figura 2.5 pentru a demonstra capabilitățile rețelei neuro-symbolice. Sunt implementați toți conectorii menționați anterior și sunt prezentate rezultatele în urma folosirii lor pe predicate, care la rândul lor folosesc constante. În definirea cuantificatorilor, se poate observa parametrul $p$ cu valori de 4 pentru cuantificatorul universal și 6 pentru cel existențial. Acest parametru are rolul de a ajusta atenția asupra abaterilor din datele asupra cărora este aplicat cuantificatorul. Pentru $\forall$ avem următoarea formulă: $$ X(a_1, a_2, a_3, ..., a_{k-1}, a_k) = 1 - \bigg(\frac{1}{k}\sum_{i=1}^{k}(1 - a_i)^p\bigg)^{\frac{1}{p}} $$
În timpul antrenării, pentru a îmbunătăți abilitatea modelului de a generaliza, a fost folosită o valoare mai mică pentru $p$, în cele mai multe cazuri de 2. În schimb, pentru partea de interogări, unde formulele sunt mai simple și clar definite, au fost folosite valori mai mari precum în exemplul din Figura 2.5. Un exemplu de folosire a parametrului $p=2$, ar rezulta într-o funcție de agregare care are ca rezultat abaterea standard a valorilor oferite, iar pentru $p= \infty$ vom obține valoarea minimă.

\newpage{}

\newpage{}

\chapter{Aplicația - concept și implementare}

\section{Structura internă}
Pentru a face implementarea algoritmilor cât mai lizibilă, am decis să creez pentru fiecare algoritm un document separat cu ajutorul Jupyter Notebook \cite{Kluyver2016jupyter}. Pentru fiecare implementare a fost realizată o explorare a setului de date folosit și o prelucrare a acestuia pentru a fi compatibil.
\begin{itemize}

    \item \textbf{Primul experiment} constă în folosirea părții de acuratețe interactivă și a raționa-mentului deductiv prin implementarea unor funcții $\phi$ de interogare a setului de date KDD99 \cite{kdd99}. 
    \item \textbf{Un al doilea experiment} a fost implementarea unui clasificator multiclass de detecție și diferențiere a atacurilor din setul de date KDD99 prin algoritmul neuro-symbolic Logic Tensor Network.
    \item \textbf{Al treilea experiment} realizează tot o clasificare și diferențiere a atacurilor din același set de date, însă de data aceasta a fost implementată printr-un algoritm din Învățarea Profundă.
    \item \textbf{Al patrulea experiment} face o comparație directă a clasificării tipurilor de atacuri, realizată de cei doi algoritmi, dintr-un set de date mai nou și mai complex CIC-IDS2017 \cite{sharafaldin2018toward}.
    \item \textbf{Al cincilea experiment} ne demonstrează că algoritmul neuro-symbolic poate să fie folosit și pentru a realiza o regresie liniară. În acest caz, este realizată o predicție a poziției deteriorării unei grinzi de metal. Ca set de date a fost folosit cel din articolul \cite{gillich2022beam, }, în urma unei discuții cu autorii de la Universitatea Babeș-Bolyai din Cluj-Napoca.
\end{itemize}

\subsection{Structura seturilor de date folosite}
KDD99 este un set de date folosit în IDS(intrusion detection system) și în învățarea automată. A fost folosit pentru prima dată la The Third International Knowledge Discovery and Data Mining Tools Competition și KDD-99 The Fifth International Conference on Knowledge Discovery and Data Mining. Cele două evenimente s-au ținut simultan. Sarcina participanților la acele evenimente a fost să creeze un sistem capabil să detecteze intruziunile sub forma unor atacuri și conexiunile normale.

Sarcina primită de participanți a evidențiat beneficiile unei rețele în care sunt monitorizate conexiunile și în care se pot identifica posibilele atacuri. A fost propusă o antrenare a unui model predictiv și a fost explicat modul în care s-a făcut rost de aceste date. Atacurile din KDD99 sunt de 4 tipuri, definite foarte bine și în lucrarea \cite{kendall1999database}.

Atacuri cibernetice de tip 
\textbf{DOS}(denial-of-service) sunt caracterizate prin cereri de comunicare într-un număr foarte mare asupra țintei, repetate cu scopul folosirii resurselor acesteia, blocarea oricăror altor conexiuni posibile sau oprirea indusă forțat. Blocarea acestui tip de atacuri nu este simplă, atacurile de cele mai multe ori venind de la mai multe calculatoare și fiind necesară o monitorizare constantă a conexiunilor suspecte. În setul de date, atacurile "back", "land", "neptune", "pod", "smurf" și "teardrop" intră în această categorie. 

\textbf{R2L} reprezintă atacurile în care atacatorul încearcă să facă rost de informații locale ale ținei. Modul în care aceste atacuri decurg poate să difere, însă scopul rămâne cel menționat. Atacurile pot să încerce să creeze fișiere locale țintei de tip troian, să adreseze interogări pe calculatorul țintei cu scopul obținerii parolelor locale, să "ghicească" parola prin încercări repetate sau pur și simplu să obțină controlul asupra țintei. "ftp\textunderscore write", "imap", "multihop", "phf", "spy", warezclient", "warezmaster" sunt atacuri din setul de date KDD99 care intră în această categorie.

\textbf{U2R} au loc atunci când atacatorul are deja acces la un cont din sistemul sau rețeaua țintă cu scopul preluării controlului asupra întregului sistem. Sunt exploatate greșeli din codul programelor și sunt exploatate vulnerabilitățile pentru a obține accesul la superutilizator. Chiar dacă acest lucru poate fi evitat, programele complexe sunt predispuse la aceste atacuri. Atacurile de tip U2R din setul de date sunt "buffer\textunderscore overflow", "loadmodule", "perl", "rootkit".

\textbf{probe} sunt atacurile care au ca scop aflarea posibilelor vulnerabilități ale țintei pentru un viitor atac. "ipsweep", "nmap", "portseep" și "satan" sunt astfel de atacuri. De obicei sunt țintite porturile, adresele IP asociate țintei.

Pe site-ul \cite{kdd99} sunt mai multe variante ale KDD99, dintre care și setul de date care conține doar 10\% din conexiuni, multe articole publicate considerând acest procent suficient. Totuși, în scopul obținerii unui model cât mai capabil să detecteze posibilele atacuri, s-a folosit setul de date original, întreg.

Înainte de a crea rețeaua neuronală profundă și cea neuro-symbolică, a fost realizată o explorare a acestui set de date. În Figura 3.1,  putem observa repartiția atacurilor, în ordine crescătoare a numărului conexiunilor lor: "spy", "perl", "phf", "multihop", "ftp\textunderscore write", "loadmodule", "rootkit", "imap", "warezmaster", "land", "buffer\textunderscore overflow", "guess\textunderscore passwd", "nmap", "pod", "teardrop", "warezclient", "portsweep". "ipsweep", "satan", "back", "normal", "neptune", "smurf". Datale au fost reprezentate cu o scală logaritmică, deoarece repartiția acestora nu este uniformă, fiind , de exemplu, o diferență foarte mare în numărul conexiunilor de tip "spy" și conexiuni de tip "smurf".
\begin{figure}[!htbp]
    \centering
    \includegraphics[height= 240px,width= 260px]{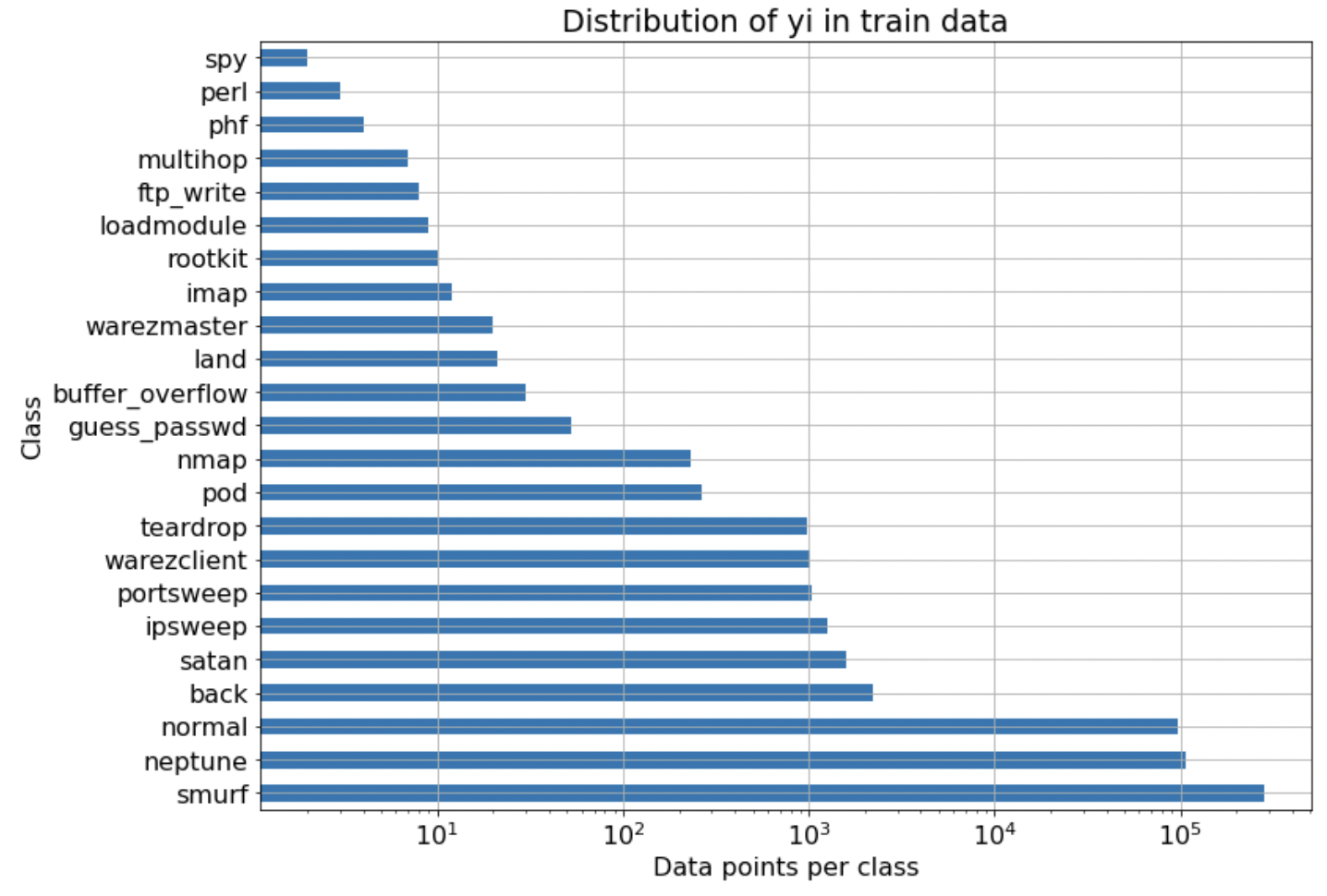}
    \caption{Before removal of duplicates}
    \label{fig:my_label}
\end{figure}

Pentru a încerca să rezolvăm această problema a uniformității repartiției datelor, au fost grupate atacurile în funcție de categoria din care fac parte(categoria de conexiuni normale și DoS, R2L, U2R, probe, descrise anterior). Astfel, din 23 de tipuri diferite de conexiuni, acum vom avea 5 categorii. În Figura 3.2 se poate observa repartiția datelor după gruparea conexiunilor pe categorii. Și de această dată a fost aleasă o scală logaritmică pentru reprezentare. Această grupare a datelor are rolul și de a face totuși posibilă antrenarea unui model, în stadiul inițial al setului de date fiind o diferență prea mare în numărul de conexiuni ale unor atacuri.

\begin{figure}[!htbp]
    \centering
    \includegraphics[height= 170px,width= 220px]{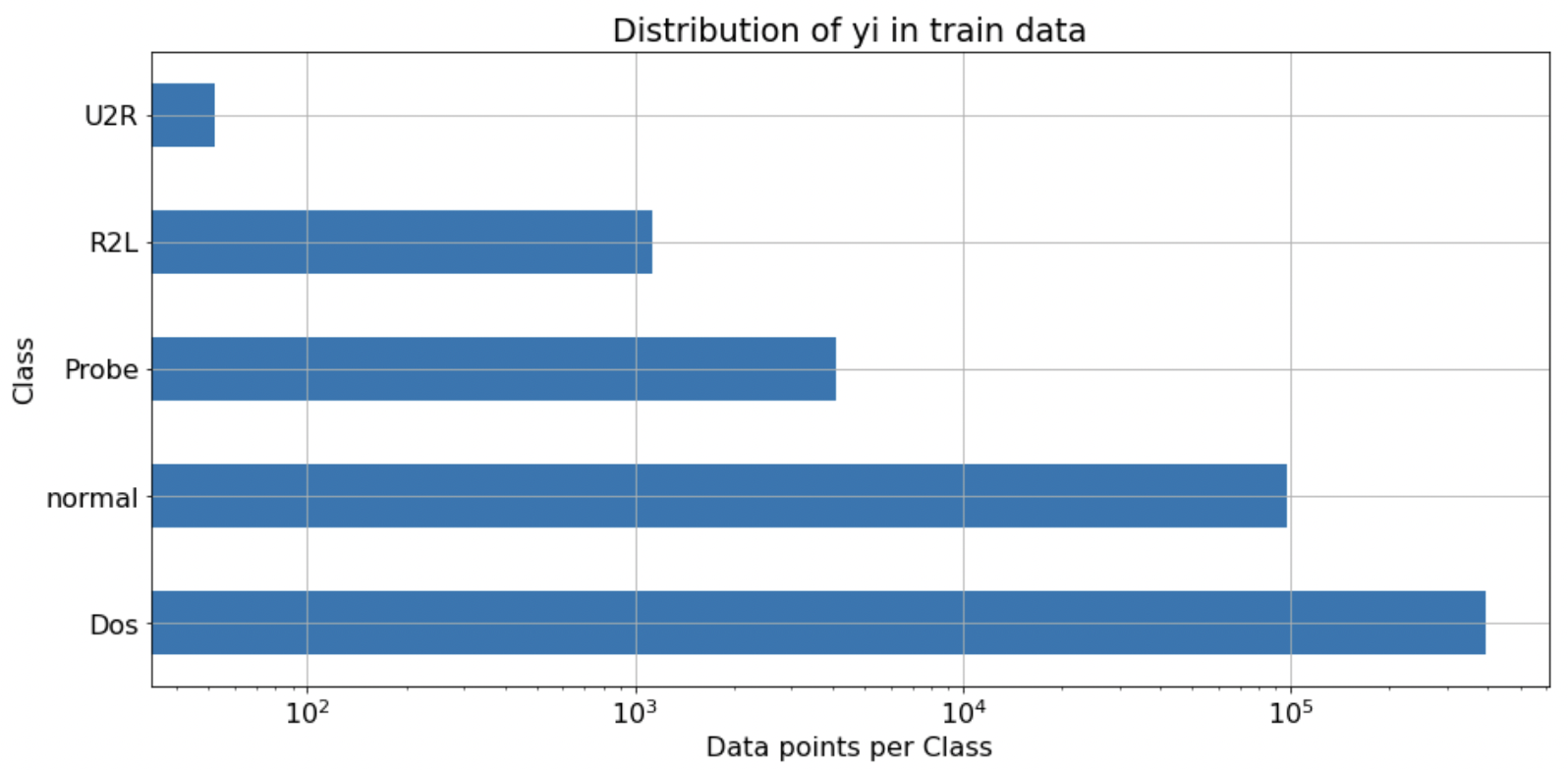}
    \caption{Before removal of duplicates}
    \label{fig:my_label}
\end{figure}

Am continuat prin a face o curățare a setului de date rezultat în urma grupării conexiunilor, prin eliminarea 
conexiunilor dublate și a eliminării posibilelor valori de NaN. Rezultatele se pot observa în Figura 3.3. De remarcat este că, din cauza modului în care funcționează atacurile "Dos", prin numărul ridicat de cereri de comunicare asupra țintei, eliminarea liniilor duplicate din setul de date face ca de această dată cele mai multe conexiuni să fie de tipul "normal".

\begin{figure}[!htbp]
    \centering
    \includegraphics[height= 170px, width= 220px]{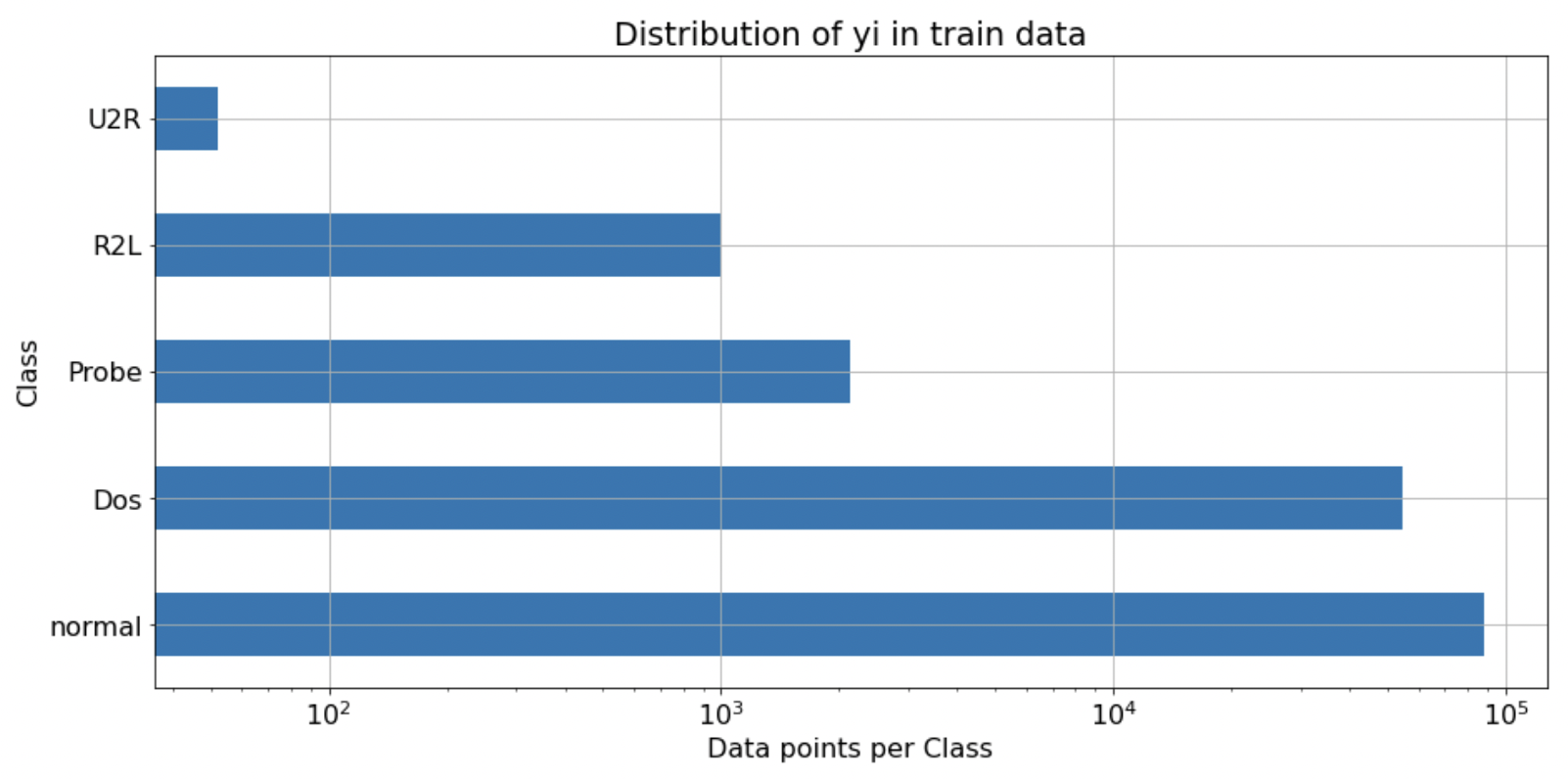}
    \caption{After removal of duplicates}
    \label{fig:my_label}
\end{figure}

Pentru experimentul 4, s-a folosit setul de date \textbf{CIC-IDS2017}. Am decis necesară experimentarea și pe un set de date actual și mai complex folosit în antrenarea sistemelor de detecție a intruziunilor în traficul de internet, întrucât mai multe articole \cite{al2018kdd}, \cite{KDD99harmful}, \cite{thapa2021secure} critică setul de date KDD99 deoarece acesta este vechi, prezintă inconsistențe în conexiuni și este o falsă reprezentare a traficului unei rețele reale aflată sub atac. Concret, acest set de date aduce aproximativ de două ori mai multe atribute pentru fiecare conexiuni, considerate mult mai relevante de către autori. Relevanța stă prin urmarea urmatoarelor criterii în crearea setului de date: folosirea unei configurații de internet care cuprinde întreaga topologie de rețea, înregistrarea întreagă a traficului de internet din rețea, secționarea conexiunilor pe zile și pe tipul acestora, folosirea celor mai comune protocoale de internet, folosirea celor mai comune atacuri raportate de McAfee în 2016 precum și alte criterii legate de modul în care au fost înregistrate aceste conexiuni.
Setul de date monitorizează activitatea de internet pe o perioadă de 5 zile și arată în felul următor. 
\begin{itemize}
    \item Luni, 3 iulie, 2017 - au fost inregistrate doar activităti normale, nefiind prezente atacuri. După împărțirea pe zile, setul de date din această zi care include conexiunile normale are dimensiunea de 11GB.
    
    \item Marti, 4 iulie, 2017 - atacuri Brute Force precum FTP-Patator și SSH-Patator au avut loc, fiecare, în decurs de câte o oră. În restul zilei, au fost înregistrate activitățile normale. Dimensiunea acestei secțiuni a setului de date este tot de 11GB.
    
    \item Miercuri, 5 iulie, 2017 - au avut loc atacuri DoS/DDoS precum DoS slowloris, DoS Slowhttptest, DoS Hulk, DoS GoldenEye, și atacuri Heartbleed. Au fost înregistrate aceste atacuri, împreună cu conexiunile normale din restul zilei. Dimensiunea este de 13GB.
    
    \item Joi, 6 iulie, 2017 - atacuri Brute Force, XSS, SQL Injection, PortScan și exploatări ale sistemelor de operare au fost înregistrate. Desigur, activitatea normală a fost și de această dată înregistrată. Dimensiunea este de 7.8GB. 
    
    \item Vineri, 7 iulie, 2017 - în această zi au avut loc atacuri Botnet și pe finalul zilei PortScan și DDoS. În experimentul 4, s-a folosit clasificarea pe a doua parte a zilei acesteia, cea cu atacurile PortScan și DDos. Atacurile acestea, împreună cu activitatea normală, au 8.3GB.
\end{itemize} 

Prin combinarea conexiunilor de tup DDoS și PortScan din ziua de vineri, precum și a restului activitatii normale, s-a obținut un set de date cu o repartiție a conexiunilor precum în figura 3.4. Numărul de atacuri DDoS fiind de 212718, pentru atacurile PortScan 128005, iar pentru conexiunile normale înregistrandu-se 57305.

\begin{figure}[!htbp]
    \centering
    \includegraphics[width=0.6\textwidth]{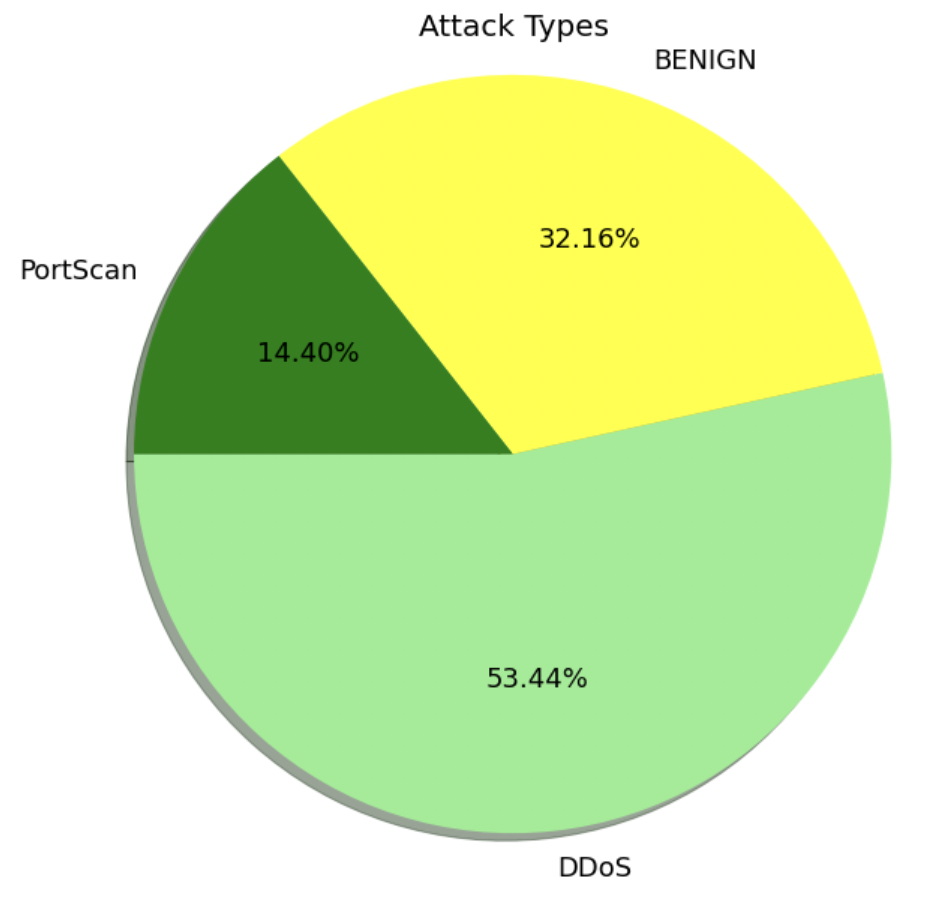}
    \caption{Repartiția atacurilor CIC-IDS2017}
    \label{fig:my_label}
\end{figure}

Pentru ultimul experiment, a fost folosit setul de date din articolul \cite{gillich2022beam} de predicție a poziției deteriorării unei grinzi. În acest set de date, valorile sunt împărțire în două categorii. Input - 8 valori RFS ale vibrațiilor și Target - unde avem severitatea defectului transversal al grinzii, poziția defectului și severitatea încastrării.

\subsection{Cazurile de utilizare}
\begin{figure}[!htbp]
    \centering
    \hspace*{-1cm} 
    \includegraphics[width=1.1\textwidth]{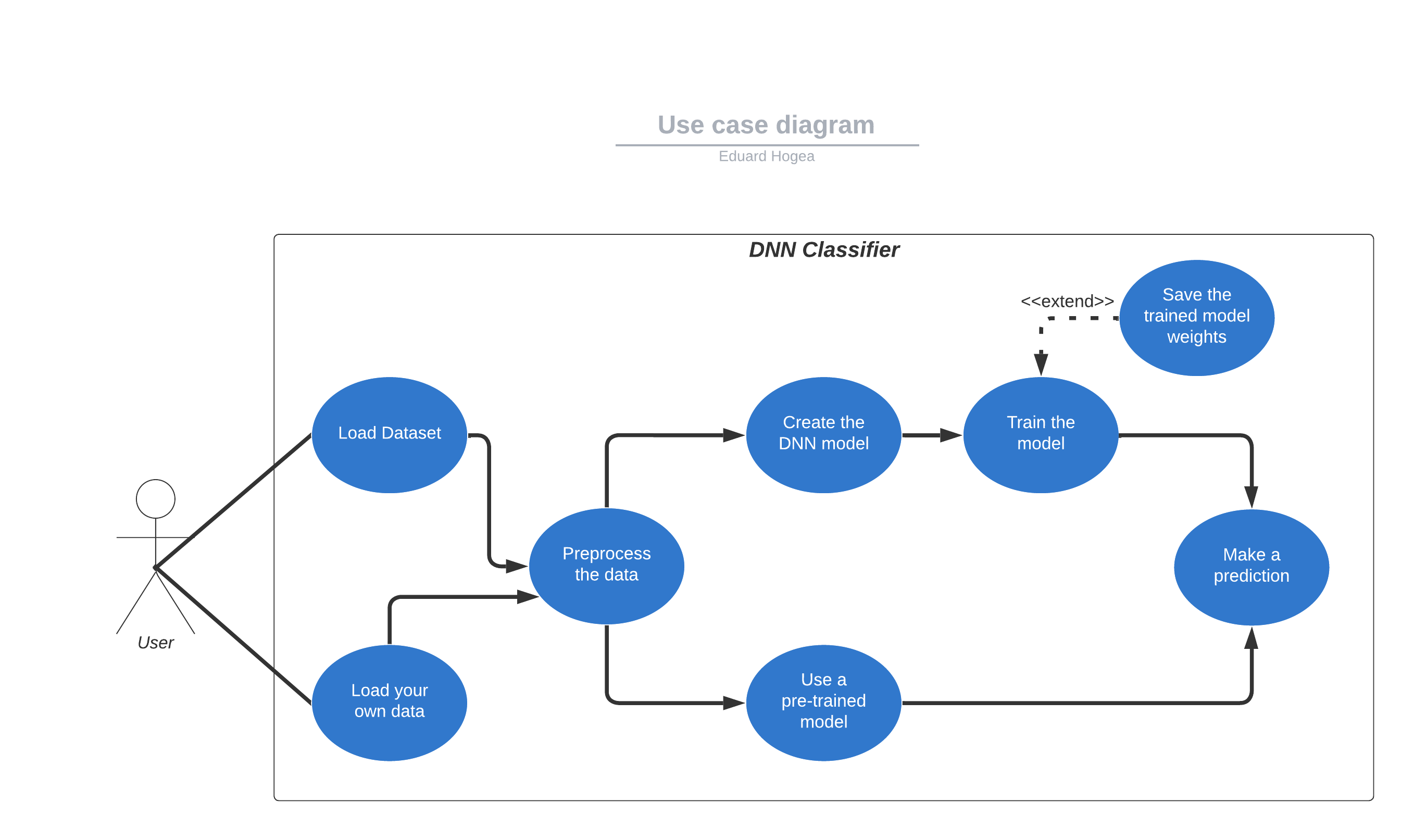}
    \caption{Cazul de utilizare pentru un clasificator DNN}
    \label{fig:my_label}
\end{figure}

\begin{figure}[!htbp]
    \hspace*{-2cm} 
    \includegraphics[width=1.25\textwidth]{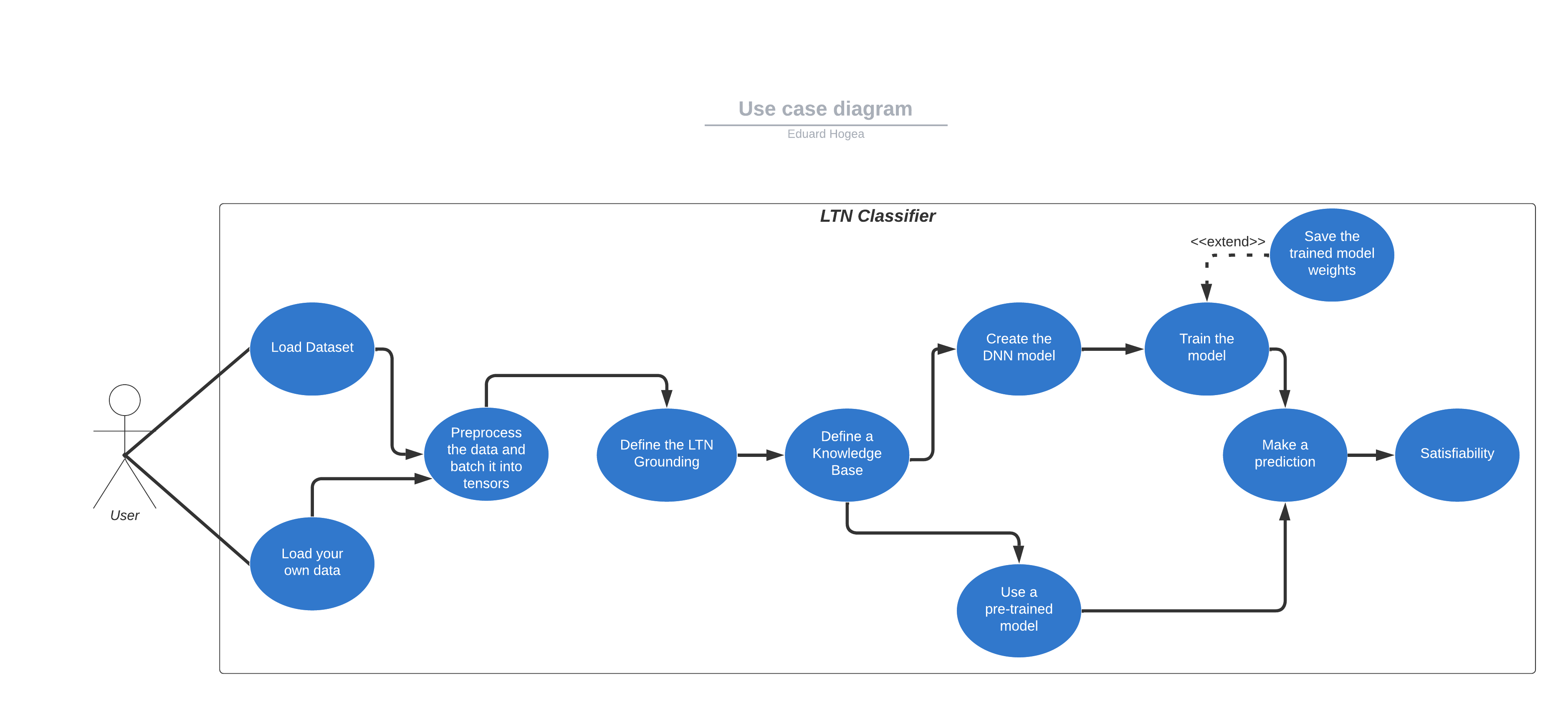}
    \caption{Cazul de utilizare pentru un clasificator LTN}
    \label{fig:my_label}
\end{figure}

Pentru a ilustra cum funcționează clasificarea realizată de modelul de Învățare Profundă și cea realizată de modelul neuro-symbolic, am realizat cazurile de utilizare din figurile 3.1 și 3.2. Pentru cazul de utilizare al rețelei neuronale profunde un utilizator are două posibilități. Prima dintre ele, atunci când utilizatorul dorește să folosească un model deja antrenat și să obțină o predicție pe datele furnizate. În acest mod, nu este necesară crearea unui model al rețelei și antrenarea acestuia, și se poate folosi direct un model antrenat anterior prin încărcarea valorilor ponderilor dintre neuroni pentru a obține o predicție. Al doilea mod în care utilizatorul poate să folosească acest sistem de clasificare este prin încărcarea unui set de date și, desigur, crearea și antrenarea unui model. 





\subsection{Diagrama de activitate}
\begin{figure}[!htbp]
    \hspace*{-0.5cm} 
    \includegraphics[width=1\textwidth]{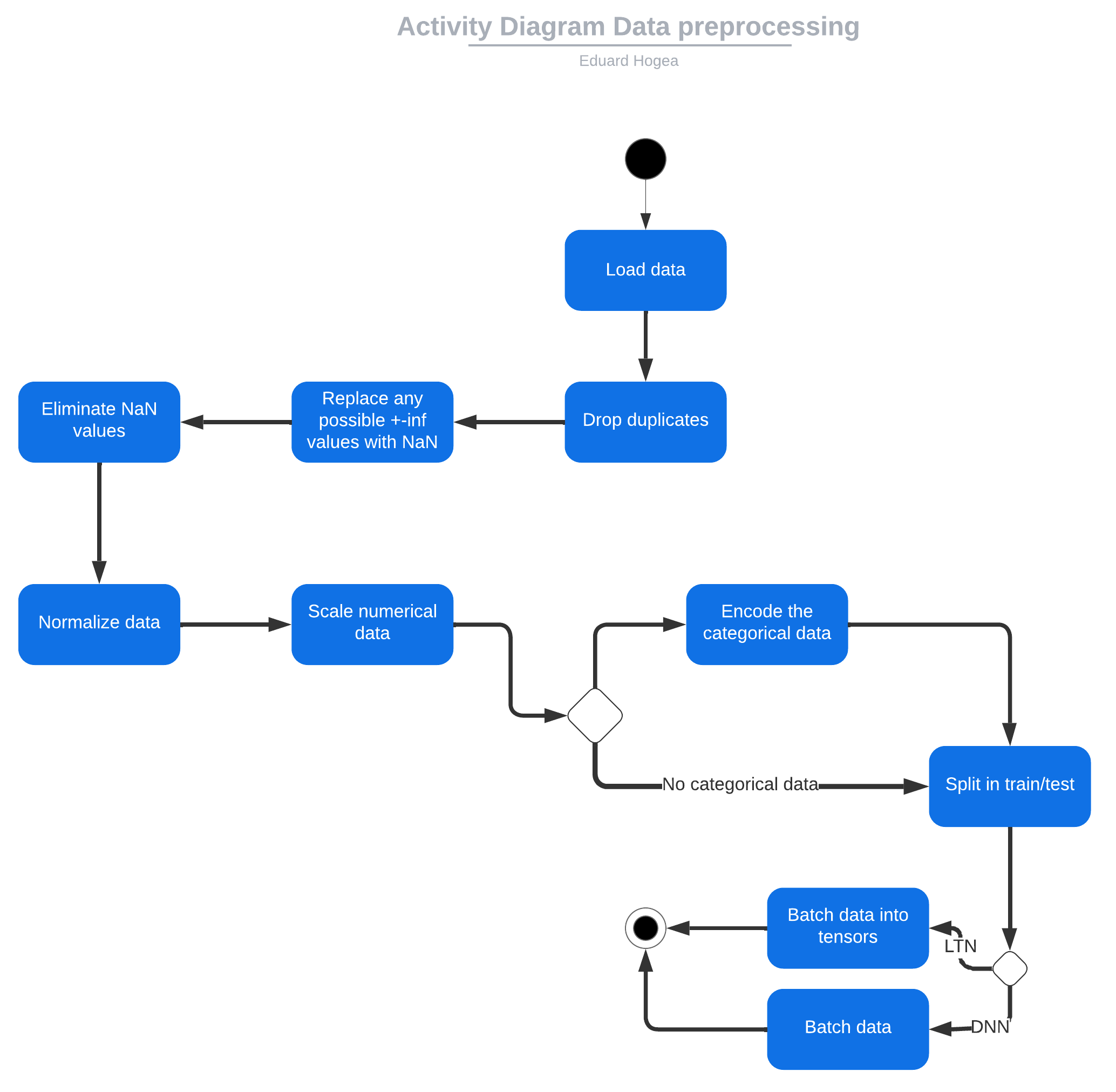}
    \caption{Diagrama de activitate pentru preprocesarea datelor}
    \label{fig:my_label}
\end{figure}

Diagramă de activitate pentru preprocesarea datelor din clasificări, Figura 3.3, are rolul de a explica prelucrările și modul în care ajung datele inițiale să fie transformare trimise apoi în antrenarea rețelei neuronale(grupate în tensori în cazul LTN). Este realizată o curățare a datelor prin ștergerea rândurilor duplicate și este realizată și o transformare, urmată de o eliminare a intrărilor care au în componentă valori de $+-\infty$ sau $NaN$. Unde a fost cazul, este realizată o grupare a conexiunilor și este realizată o împărțire a datelor numerice și categorice. Datele numerice sunt scalate iar cele categorice sunt encodate corespunzător. În final, rezultatele sunt împărțite în setul de antrenare și cel de testare, acesta urmând să fie trimise mai departe. În cazul setului de date pentru detectarea pozitiei unui defect într-o grindă, datele nu au mai fost scalate, scopul fiind vederea unei comparații clare între locul prezis al defectului și cel actual.


\section{Dependențele necesare în rularea experimentelor}
În rularea experimentelor, au fost folosite următoarele biblioteci:
\begin{enumerate}
    \item Python - 3.9.7
    \item Tensorflow-macos - 2.7.0
    \item Keras - 2.7.0
    \item Pandas - 1.3.4
    \item Numpy - 1.21.4
    \item Matplotlib - 3.5.0
\end{enumerate}

Ca mediu de dezvoltare a fost folosit Jupyter. Toate acestea au fost instalate folosind Anaconda și programul de instalare a pachetelor pip, iar rezultatele obținute se datorează rulării experimentelor pe următoarea configurație: Macbook Air M1(2020), Apple M1 Chip, 8Gb Memory, macOS Montrey 12.3.

A fost instalată platforma Anaconda, folosită pentru gestionarea pachetelor. Aceasta vine implicit cu următoarele dependețe folosite pre-instalate: Python, matplotlib, numpy. A fost creat un mediu virtual, unde au fost instalate și restul bibliotecilor. O excepție a fost făcută în instalarea Logic Tensor Networks, unde a fost nevoie de o clonare a versiunii de pe github \cite{versiuneltn} și o instalare locală a acestuia în mediul creat.

\newpage{}

\chapter{Experimente numerice}
Experimentele numerice din această lucrare au scopul de a justifica nevoia dezvoltării sistemelor hibride și de a arăta rezultatele care pot fi obținute pe seturile de date KDD99, CIC-IDS2017 și cel de detecție a defectelor descrise în capitolul anterior. Ca un reper în comparație, pentru aceste seturi de date au fost implementați și algoritmi de clasificare și de regresie bazați pe arhitecturi de Învățare Profundă.
\section{Clasificarea bazată pe protocolul folosit și statusul conexiunii KDD99}
Primul experiment realizat este cel în care s-a folosit algoritmul LTN pentru a clasifica conexiunile din setul de date KDD99 prelucrat cu modificările detaliate în capitolul 3, cu partea de Real Logic definită cu ajutorul tipului de conexiune și statusul acesteia. Scopul acestui experiment este acela de a înțelege relațiile dintre atributele conexiunilor și de a interpreta valorile din setul de date, precum și acela de a demonstra un alt posibil caz de utilizare al sistemelor hibride, acela de a realiza interogări complexe și de a avea acuratețe interactivă și un raționament deductiv. Stratul de intrare al rețelei este format din tipurile de protocoale și statusurile de conexiuni, precum în tabelul de mai jos.

\begin{table}[!htp]
\begin{tabular}{|l|l|l|l|l}
\hline
0.TCP   & 1.ICMP    & 2.UDP  & 3.SF  & \multicolumn{1}{l|}{4.S1}   \\ \hline
5.REJ   & 6.S2      & 7.S0   & 8.S3  & \multicolumn{1}{l|}{9.RSTO} \\ \hline
10.RSTR & 11.RSTOS0 & 12.OTH & 13.SH &                             \\ \cline{1-4}
\end{tabular}
\end{table}

Baza de cunoștințe a fost creată în așa fel încât să ne ofere informații tipurile de protocoale folosite și doar o parte din statusurile conexiunilor. Cu toate ca informațiile din baza de cunoștințe nu sunt complete, modelul este totuși capabil să realizeze clasificarea prin restul informațiilor dobândite din antrenarea prin partea de Rețea Neuronală Profundă.

Ca set de axiome, avem următorul set definit:
\begin{center}
    
\begin{description}
\centering
    \item $\forall x_{tcp}: P(x_{tcp}, tcp)$
    \item $\forall x_{icmp}: P(x_{icmp}, icmp)$
    \item $\forall x_{udp}: P(x_{udp}, udp)$
    \item $\forall x_{sf}: P(x_{sf}, sf)$
    \item $\forall x_{s1}: P(x_{s1}, s1)$
    \item $\forall x_{rej}: P(x_{rej}, rej)$
    \item $\forall x: \neg ( P(x, tcp) \land P(x, sf) )$
    \item $\forall x: \neg ( P(x, icmp) \land P(x, s1) )$
    \item $\forall x: \neg ( P(x, udp) \land P(x, rej) )$

\end{description}
\end{center}

În antrenarea modelului, sunt urmărite mai multe metrici. Au fost măsurate:

\begin{enumerate}
    \item Nivelul de satistisfiabiliate al bazei de cunoștințe pentru datele din antrenare.
    \item Nivelul de satistisfiabiliate al bazei de cunoștințe pentru datele din testare.
    \item Acuratețea pe datele din setul de antrenare (a fost folosită funcția de loss Hamming).
    \item Acuratețea pe datele din setul de testare (tot funcția de loss Hamming a fost folosită).
    \item Nivelul de satisfiabilitate al unei formule $\phi1$ interogate de-a lungul epocilor în timpul antrenării $\forall x: p(x,udp)\to \neg p(x,tcp)$ (fiecare conexiune $udp$ nu poate să fie una $tcp$ și vice-versa)
    \item Nivelul de satisfiablitate al unei formule $\phi2$ interogate de-a lungul epocilor în timpul antrenării care ne așteptăm să fie falsă $\forall x: p(x,udp) \to p(x,tcp)$ (fiecare conexiune $udp$ este și o conexiune $tcp$ one și vice-versa)
    \item Nivelul de satisfiablitate al unei formule $\phi3$ interogate de-a lungul epocilor în timpul antrenării care ne așteptăm să fie falsă. $\forall x: p(x,tcp) \to p(x,sf)$ (fiecare conexiune $tcp$ are un status de conexiune $sf$)
\end{enumerate}

Rezultatele experimentului au fost salvate într-un fișier $.csv$ care conține toate metricile enumerate mai sus. Formulele interogate $\phi1$, $\phi3$ au un nivel mare de satisfiabilitate, aproape perfect. Formula $\phi2$ este opusul primei formule, satisfiabilitatea acesteia fiind egală cu $1 - sat(\phi1)$. Din asta, se poate spune cu siguranță că o conexiune $udp$ nu poate să fie o conexiune $tcp$ și că o conexiune $tcp$ este foarte probabil să aiba un status al conexiunii $sf$. Cu toate că acest experiment oferă informații utile, acuratețea clasificării este însă slabă, baza de cunoștințe având nevoie de modificări.

\section{Clasificarea conexiunilor prin LTN KDD99}
În cel de al doilea experiment a fost realizată o noua clasificare a atacurilor din setul de date KDD99, cu prelucrările datelor explicate anterior. Într-o situație ideală, un model de detecție al intruziunilor este capabil nu doar să diferențieze între o conexiune normala și un atac, ci să recunoască și ce tip de atac este pentru a se putea lua măsuri în prevenirea sau combaterea acestuia.

Un articol foarte bun în care este realizată o analiză detaliată a setului de date KDD99 este \cite{kayacik2005selecting}. Acolo este prezentată și o comparație a impactului pe care îl au atributele conexiunilor și corelația acestora cu tipul conexiunilor. Am ținut cont de datele analizate în articolul menționat și am pastrat doar atributele relevante în realizarea clsificării. Am definit setul de axiome cu reguli clare pentru fiecare tip de atac și am adăugat constrângeri pentru a ajuta modelul să generalizeze. Baza de cunoștințe pentru acest experiment arată în felul următor:
\begin{center}

\begin{description}
\centering
    \item $\forall x_{normal}: P(x_{normal}, normal)$
    \item $\forall x_{DOS}: P(x_{DOS}, DOS)$
    \item $\forall x_{probe}: P(x_{probe}, probe)$
    \item $\forall x_{R2L}: P(x_{R2L}, R2L)$
    \item $\forall x_{U2R}: P(x_{U2R}, U2R)$
    
    
    \item $\forall x: \neg ( P(x, normal) \land P(x, DOS) )$
    \item $\forall x: \neg ( P(x, normal) \land P(x, probe) )$
    \item $\forall x: \neg ( P(x, normal) \land P(x, R2L) )$
    \item $\forall x: \neg ( P(x, normal) \land P(x, U2R) )$
    \item $\forall x: \neg ( P(x, DOS) \land P(x, probe) )$
    \item $\forall x: \neg ( P(x, DOS) \land P(x, R2L) )$
    \item $\forall x: \neg ( P(x, DOS) \land P(x, U2R) )$
    \item $\forall x: \neg ( P(x, R2L) \land P(x, probe) )$
    \item $\forall x: \neg ( P(x, R2L) \land P(x, U2R) )$
    \item $\forall x: \neg ( P(x, probe) \land P(x, U2R) )$

    \end{description}
\end{center}

O explicție pentru setul de axiome propus, în aceeași ordine, arată în felul următor:
\begin{itemize}
    \item orice non-atac trebuie să aibă atributul label normal;
    \item toate atacurile DOS trebuie să aibă label DOS;
    \item toate atacurile probe trebuie să aibă label probe;
    \item toate atacurile R2L trebuie să aibă label R2L;
    \item toate atacurile U2R trebuie sa aibă label U2R;
    \item dacă o conexiune x are label normal, ea nu poate sa aibă și label DOS;
    \item dacă o conexiune x are label normal, ea nu poate sa aibă și label probe;
    \item dacă o conexiune x are label normal, ea nu poate sa aibă și label R2L;
    \item și așa mai departe...
\end{itemize}

Fără a antrena modelul, valoarea satisfiabilității inițiale bazată doar pe logica de mai sus definită are valoarea de 0.59471. Am considerat că o antrenare a modelulul este necesară. Pentru a face comparația posibilă, metricile folosite sunt asemănătoare cu cele din experimentul anterior, cu exceptia interogării formulelor $\phi$, acestea fiind redefinite în felul următor:

\begin{enumerate}
    \item Nivelul de satisfiabilitate al unei formule $\phi1$ care ne așteptăm să fie adevărată $\forall x: p(x,normal) \to \neg p(x,DOS) $ (fiecare conexiune $normal$ nu poate să fie o conexiune $DOS$ și vice-versa)
    \item Nivelul de satisfiabilitate al unei formule $\phi1$ care ne așteptăm să fie falsă $\forall x: p(x,normal) \to p(x,probe)$ (fiecare conexiune $normal$ este și o conexiune $probe$ și vice-versa)
    \item Nivelul de satisfiabilitate al unei formule $\phi1$ care ne așteptăm să fie falsă $\forall x:  p(x,normal) \to p(x,normal)$ (fiecare conexiune $smurf$ este și o conexiune $normal$ și vice-versa)
\end{enumerate}   

Măsurătorile din timpul antrenării au fost salvate într-un fișier $.csv$ care este și reprezentat în Figura 4.1 și poate fi interpretată astfel. În prima epocă a antrenării, nivelul de satisfiabilitate pentru baza de cunoștinte a crescut de la valoarea de 0.54725 la 0.6759 pentru datele de antrenare și până la 0.7216 pentru datele de testare. Aceste rezultate de la începutul antrenării pot părea atipice, având un nivel de acuratețe mai mare pentru datele de testare, dar se datorează faptului că distribuția atacurilor în setul de date este una neuniformă.

Nivelul de satisfiablitate al unui model neuro-symbolic de acest tip este similar cu o funcție loss din Învățarea Profundă și evoluția poate fi percepută ca o optimizare a modelului asupra căruia au fost aplicate constrângeri ale logicii de ordinul întâi. Întrucât nivelul de satisfiabilitate al bazei de cunoștințe a crescut, a fost îmbunătațită și acuratețea modelului, cu valori plecând de la 0.7757 pentru datele de antrenare și 0.9162 pentru datele de testare în primele epoci și ajungând la 0.9954 și respectiv 0.9947 în ultimele epoci ale antrenării. Comparabil, acuratețea pentru rețeaua standard de Învățare Profunda a atins valori de 0.98863 pentru datele de antrenare și 0.9869 pentru datele de testare(discutate mai mult în următorul experiment).

\begin{figure}[!htbp]
\centering
\includegraphics[height= 220px, width= 251px]{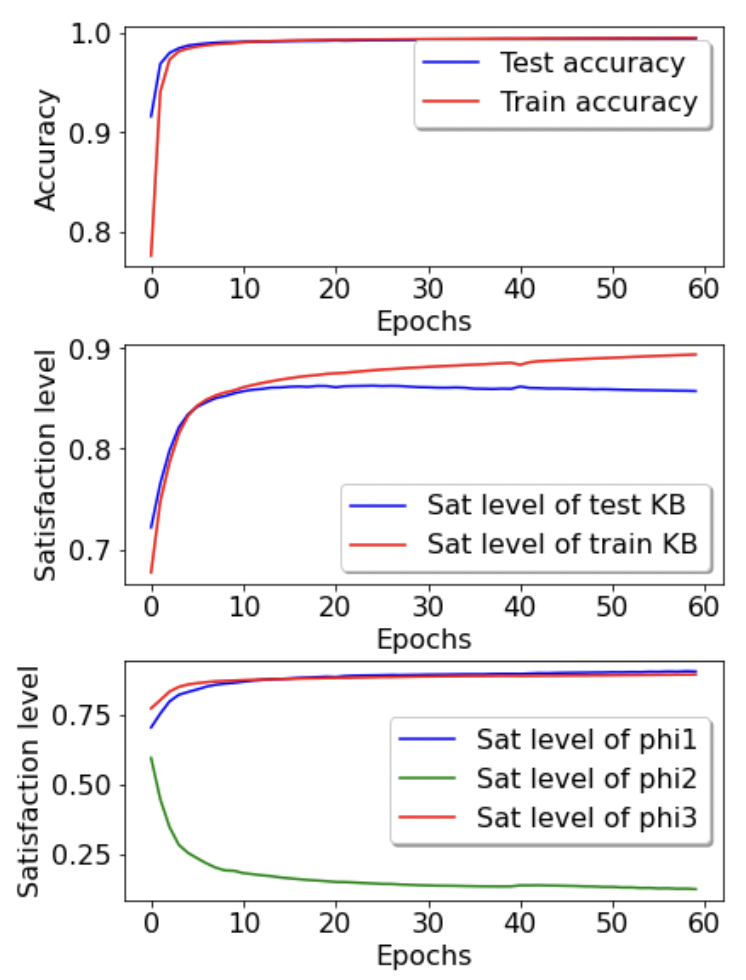}
\caption{Comparison of Accuracy, Satisfiability and Querying of Constraints}
\label{fig:my_label}
\end{figure}

\section{Clasificarea conexiunilor prin rețea neuronală profundă KDD99}
Același experiment prezentat anterior a fost reluat, însă de această dată fară partea simbolică. Din această cauză, raționamentul asupra datelor este realizat doar prin nivelul de acuratețe, facând astfel raționamentul deductiv, acuratețea interactivă și abilitatea de a aplica constrângeri prin Real Logic imposibile. De dragul consistenței, în acest experiment, următoarele aspecte au fost păstrate ca în implementarea LTN. Normalizarea datelor, encodarea one-hot a celor categorice, numărul de straturi ascunse și numărul de neuroni de pe fiecare strat ascuns au rămas identice. Totuși, diferențe notabile între cele doua implementari sunt în modul în care datele au fost trimise în rețeaua neuronală, în cazul LTN fiind stocate în tensori urmând ca aceștia să fie trimși în loturi(batch-uri); pentru ultimul strat, rețeata LTN are o funcție de activare $sigmoid$, iar rețeaua de Învățare Profundă are $softmax$; iar în ultimul rând, straturile ascunse din cazul neuro-symbolic au $elu$ ca funcție de activare, contrar funcției $relu$.

După cum se poate observa în Figura 4.2, rezultatele modelului de detecție a intruziunilor sunt foarte bune. Cu valori de acuratețe aproape perfect, 0.98863 pentru datele din antrenare și 0.9869 pentru datele de validare.

    \begin{figure}[!htbp]
    \centering
    \includegraphics[height= 230px, width= 251px]{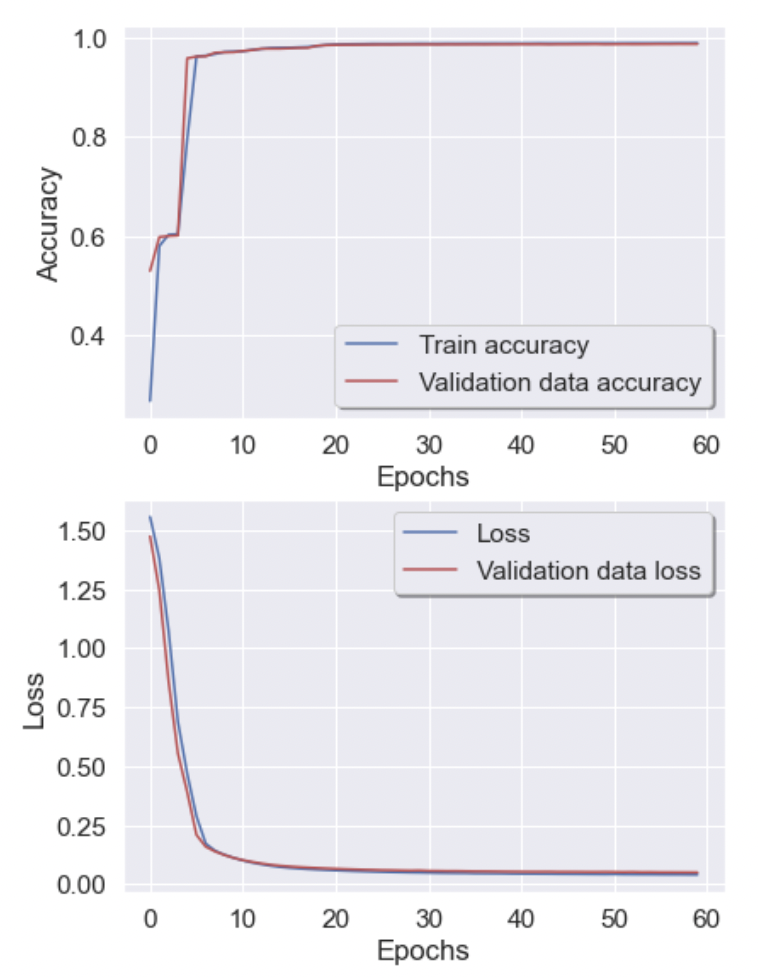}
    \caption{Evoluția loss și a acurateței din antrenare}
    \label{fig:my_label}
    \end{figure}

\section{Comparație clasificare neruonală profundă și neuro-symbolică pentru CIC-IDS2017}
În acest experiment a fost folosit setul de date CIC-IDS2017, cu diferențele dintre acest set de date și KDD99 discutate în capitolul anterior al acestei lucrări. Pentru a arăta capabilitățile LTN, am realizat o clasificare multiclass dar de această dată, single-label contrar multi-label din experimentele trecute.

Pentru modelul neuro-symbolic, datele au fost stocate în tensori și a fost creat un set de axiome. Baza de cunoștințe în acest caz este una mai simplă, am definit doar că atacurile DDoS trebuie să aibă label "DDoS", toate conexiunile BENIGN să aibă label "BENIGN" și că toate cele "PortScan" să aibă un label "PortScan". Nu a fost nevoie de aplicarea constrângerilor, deoarece numarul de clase a fost mic. Funcția din LTN implementată prin Real Logic a rămas aceeași(o rețea neuronală profundă fără schimbări asupra straturilor ascunse), precum și metricile măsurate, excepția fiind scoaterea interogărilor formulelor $\phi$.

În Figura 4.3, se poate observa o evoluție clară a acurateții modelului LTN, dar este loc de îmbunătațire. Am atins o acuratețe de 0.950, cu nivele de satisfiabilitate apropiate. Cele doua grafice sunt foarte similare, după cum este de așteptat în cazul unei baze de cunoștinte precum cea din acest experiment. 

    \begin{figure}[!htbp]
    \centering
    \includegraphics[height= 240px, width= 251px]{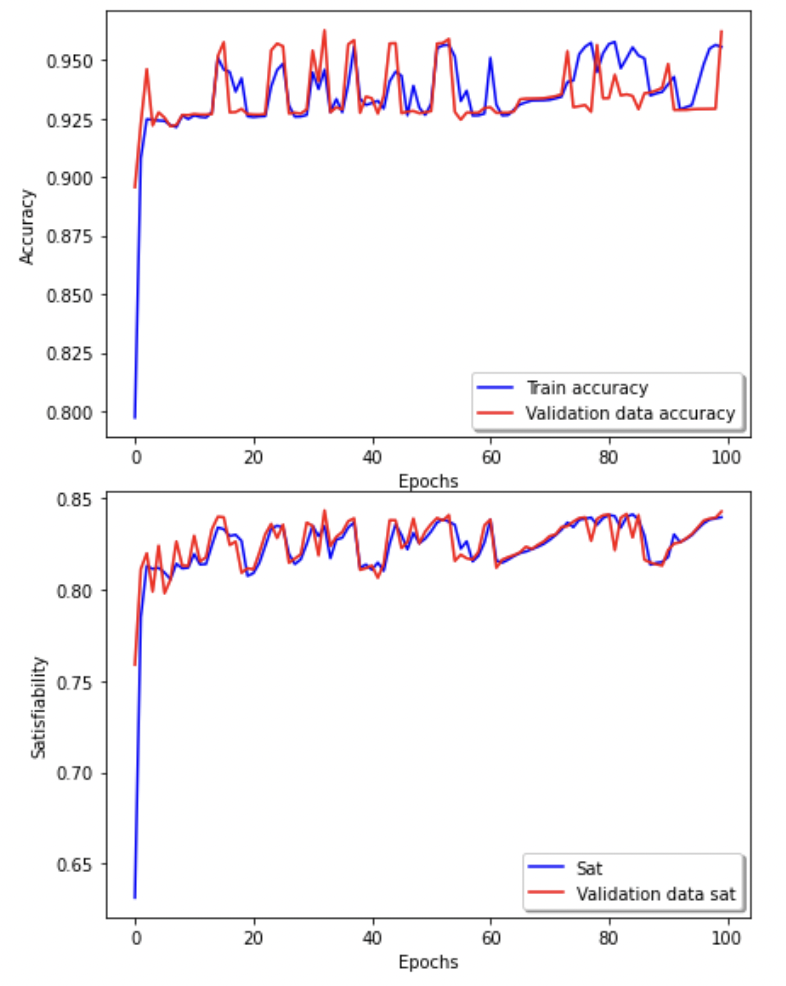}
    \caption{Evoluția satisfiabilității și a acurateței modelului LTN în antrenare}
    \label{fig:my_label}
    \end{figure}

Pentru abordarea de învățare neuronală profundă, rețeaua implementată este aproape identică cu cea folosită anterior în funcția din LTN. Rezultatele pot fi observate în Figura 4.4 și trebuie remarcat următorul aspect. Cu toate că acuratețea antrenării este într-o continua creștere, acuratețea datelor de validare este una instabilă. O comparație directa a celor doua implementări arată ca modelul LTN are rezultate mai stabile, mai constante și este mai bun în generalizarea pe date noi, cu ajutorul bazei de cunoștințe.   
    
    \begin{figure}[!htbp]
    \centering
    \includegraphics[height= 220px, width= 251px]{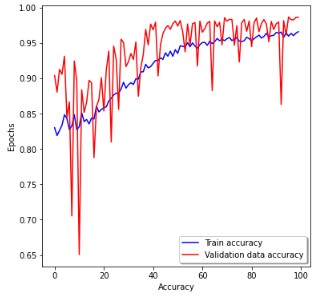}
    \caption{Evoluția acurateței în antrenare a modelului DNN}
    \label{fig:my_label}
    \end{figure}

\section{Regresie liniară neuro-symbolică pentru detecția defectelor}
Ultimul experiment al acestei lucrări realizează o regresie liniară pe setul de date cu diferite scenarii de deteriorare a unei grinzi. Scopul regresiei este acela de a prezice locația defectului, prin oferirea de date de intrare a valorilor vibrațiilor RFS(relative frequency shifts) descrise în articolele \cite{gillich2022beam, gillich2020calculus, gillich2011correlate}. Sistemul hibrid de aceasta data este format din o funcție din Real Logic care reprezintă o rețea neuronală profundă și 3 predicate de măsurare a similarității/distanței dintre poziția prezisă de rețea și poziția actuală. Aceste predicate aplică constrângeri în parametrii funcției, iar în acest experiment au fost alese urmatoarele:

\textbf{Distanța/Similaritatea Euclidiană}
$$D(x,y) = \sqrt{(\sum_{i=1}^{n}(x_i-y_1)^2)}$$
\textbf{Distanța/Similaritatea Manhattan}
$$D(x,y) = \sum_{i=1}^{k}|x_i-y_1|$$
\textbf{Distanța/Similaritatea Minkowski}
$$D(x,y) =  (\sum_{i=1}^{n}|x_i-y_1|^p)^{\frac{1}{p}}$$
Cu parametrul $p$ care poate avea diferite valori. O valoare de $p=1$ va produce rezultate similare cu distanța Manhattan, o valoare de $p=2$ cu distanța Euclidiana și o valoare de $p = \infty$ va produce valori similare cu distanța Chebyshev.

În antrenare, au fost aplicate interogări constante asupra nivelului de satisfiablitate al bazei de cunoștințe pentru datele de antrenare și cele de testare. Pentru a obține acuratețea rețelei, a fost folosită funcția RMSE(Root mean squared error) dar este important de menționat ca antrenarea retelei nu a fost afectată de această valoare, scopul antrenării fiind acela de a maximiza valoarea de satisfiablitate a axiomelor.

În urma antrenării, am obținut valori ale acurateței cuprinse între ~0.01-0.03 pentru datele de validare. Cu toate că aceste valori sunt foarte bune, în cele mai multe cazuri un set de date cu un volum așa de mare nu va fi disponibil. Din această cauză, am decis folosirea K-fold cross validation pe datele amestecate, setul de date fiind împărțit in 2 și 4 parți pentru fiecare predicat și testarea modelelor pe un nou set de validare. Setul de validare este compus din 90 de valori noi, reale, măsurate de către autorii de la Universitatea Babeș-Bolyai în mod special pentru această lucrare și una viitoare pentru jurnalul Computers in Industry publicat de catre Elsevier.

Experimentul acesta cuprinde urmatoarele:
\begin{itemize}
    \item K-fold cross validation, k=2 pentru predicatul de măsurare a distanței Euclidiene
    \item K-fold cross validation, k=4 pentru predicatul de măsurare a distanței Euclidiene
    \item K-fold cross validation, k=2 pentru predicatul de măsurare a distanței Manhattan
    \item K-fold cross validation, k=4 pentru predicatul de măsurare a distanței Manhattan
    \item K-fold cross validation, k=2 pentru predicatul de măsurare a distanței Minkowski
     
\end{itemize}

Rezultatele prezicerii poziției pentru cazurile menționate se pot vedea în următoarele tabele. În crearea acestora, am pus doar cele mai slabe 5 rezultate obținute pe datele de validare. Am notat cu $y$ valoarea poziției din setul de validare, cu $y_pred$ valoarea prezisă de catre model și cu $dif$ valoarea diferenței absolute dintre cele două poziții.

\begin{table}[!htb]
    \caption{K=2, distanță euclidiană}
    \begin{subtable}{0.1\linewidth}
      \centering
        \caption{}
        \begin{tabular}{|l|l|l|l|l|l|}
\hline
y       & 0.98  & 0.98  & 0.946 & 0.414 & 0.516 \\ \hline
y\_pred & 0.838 & 0.838 & 0.835 & 0.335 & 0.450 \\ \hline
dif     & 0.142 & 0.142 & 0.111 & 0.079 & 0.066 \\ \hline
\end{tabular}
    \end{subtable}%
    \begin{subtable}{1.45\linewidth}
      \centering
        \caption{}
\begin{tabular}{|l|l|l|l|l|l|}
\hline
y       & 0.563  & 0.56  & 0.59 & 0.516 & 0.516 \\ \hline
y\_pred & 0.507 & 0.508 & 0.542 & 0.469 & 0.469 \\ \hline
dif     & 0.055 & 0.052 & 0.047 & 0.047 & 0.046 \\ \hline
\end{tabular}
    \end{subtable} 
\end{table}

\begin{table}[!htb]
    \caption{K=4, distanță euclidiană}
    \begin{subtable}{0.1\linewidth}
      \centering
        \caption{}
        \begin{tabular}{|l|l|l|l|l|l|}
\hline
y       & 0.98  & 0.414  & 0.98 & 0.414 & 0.946 \\ \hline
y\_pred & 0.796 & 0.530 & 0.864 & 0.524 & 1.031 \\ \hline
dif     & 0.184 & 0.116 & 0.116 & 0.11 & 0.085 \\ \hline
\end{tabular}
    \end{subtable}%
    \begin{subtable}{1.45\linewidth}
      \centering
        \caption{}
\begin{tabular}{|l|l|l|l|l|l|}
\hline
y       & 0.056  & 0.081  & 0.255 & 0.687 & 0.233 \\ \hline
y\_pred & 0.004 & 0.041 & 0.224 & 0.714 & 0.205 \\ \hline
dif     & 0.052 & 0.040 & 0.031 & 0.027 & 0.027 \\ \hline
\end{tabular}
    \end{subtable} 
        \begin{subtable}{0.1\linewidth}
      \centering
        \caption{}
        \begin{tabular}{|l|l|l|l|l|l|}
\hline
y       & 0.946  & 0.946  & 0.466 & 0.98 & 0.173 \\ \hline
y\_pred & 0.900 & 0.912 & 0.493 & 0.956 & 0.151 \\ \hline
dif     & 0.045 & 0.034 & 0.027 & 0.024 & 0.022 \\ \hline
\end{tabular}
    \end{subtable}%
    \begin{subtable}{1.45\linewidth}
      \centering
        \caption{}
\begin{tabular}{|l|l|l|l|l|l|}
\hline
y       & 0.81  & 0.82  & 0.98 & 0.347 & 0.563 \\ \hline
y\_pred & 0.759 & 0.770 & 0.931 & 0.300 & 0.517 \\ \hline
dif     & 0.05 & 0.049 & 0.048 & 0.046 & 0.046 \\ \hline
\end{tabular}
    \end{subtable} 
\end{table}

\begin{table}[!htb]
    \caption{K=2, distanță manhattan}
    \begin{subtable}{0.1\linewidth}
      \centering
        \caption{}
        \begin{tabular}{|l|l|l|l|l|l|}
\hline
y       & 0.98  & 0.98  & 0.414 & 0.489 & 0.414 \\ \hline
y\_pred & 0.852 & 0.861 & 0.466 & 0.540 & 0.462 \\ \hline
dif     & 0.127 & 0.118 & 0.052 & 0.051 & 0.048 \\ \hline
\end{tabular}
    \end{subtable}%
    \begin{subtable}{1.45\linewidth}
      \centering
        \caption{}
\begin{tabular}{|l|l|l|l|l|l|}
\hline
y       & 0.98  & 0.414  & 0.98 & 0.414 & 0.489 \\ \hline
y\_pred & 0.921 & 0.458 & 0.947 & 0.446 & 0.511 \\ \hline
dif     & 0.058 & 0.044 & 0.032 & 0.032 & 0.022 \\ \hline
\end{tabular}
    \end{subtable} 
\end{table}

\begin{table}[!htb]
    \caption{K=4, distanță manhattan}
    \begin{subtable}{0.1\linewidth}
      \centering
        \caption{}
        \begin{tabular}{|l|l|l|l|l|l|}
\hline
y       & 0.414  & 0.414  & 0.59 & 0.563 & 0.414 \\ \hline
y\_pred & 0.468 & 0.441 & 0.616 & 0.54 & 0.437 \\ \hline
dif     & 0.054 & 0.027 & 0.027 & 0.023 & 0.023 \\ \hline
\end{tabular}
    \end{subtable}%
    \begin{subtable}{1.45\linewidth}
      \centering
        \caption{}
\begin{tabular}{|l|l|l|l|l|l|}
\hline
y       & 0.056  & 0.466  & 0.76 & 0.946 & 0.563 \\ \hline
y\_pred & 0.0297 & 0.487 & 0.779 & 0.964 & 0.545 \\ \hline
dif     & 0.026 & 0.021 & 0.019 & 0.018 & 0.018 \\ \hline
\end{tabular}
    \end{subtable} 
        \begin{subtable}{0.1\linewidth}
      \centering
        \caption{}
        \begin{tabular}{|l|l|l|l|l|l|}
\hline
y       & 0.98  & 0.946  & 0.946 & 0.946 & 0.056 \\ \hline
y\_pred & 0.894 & 0.869 & 0.87 & 0.877 & 0.014 \\ \hline
dif     & 0.086 & 0.077 & 0.076 & 0.069 & 0.042 \\ \hline
\end{tabular}
    \end{subtable}%
    \begin{subtable}{1.45\linewidth}
      \centering
        \caption{}
\begin{tabular}{|l|l|l|l|l|l|}
\hline
y       & 0.98  & 0.98  & 0.489 & 0.414 & 0.687 \\ \hline
y\_pred & 0.92 & 0.935 & 0.456 & 0.445 & 0.711 \\ \hline
dif     & 0.06 & 0.045 & 0.033 & 0.031 & 0.024 \\ \hline
\end{tabular}
    \end{subtable} 
\end{table}

\newpage

O reprezentare mai clară a rezultatelor pe întreg setul de date de validare, pentru fiecare scenariu se poate vedea în graficele următoare.

\begin{figure}
\centering
\begin{subfigure}{0.9\textwidth}
  \centering
  \includegraphics[width=0.95\linewidth]{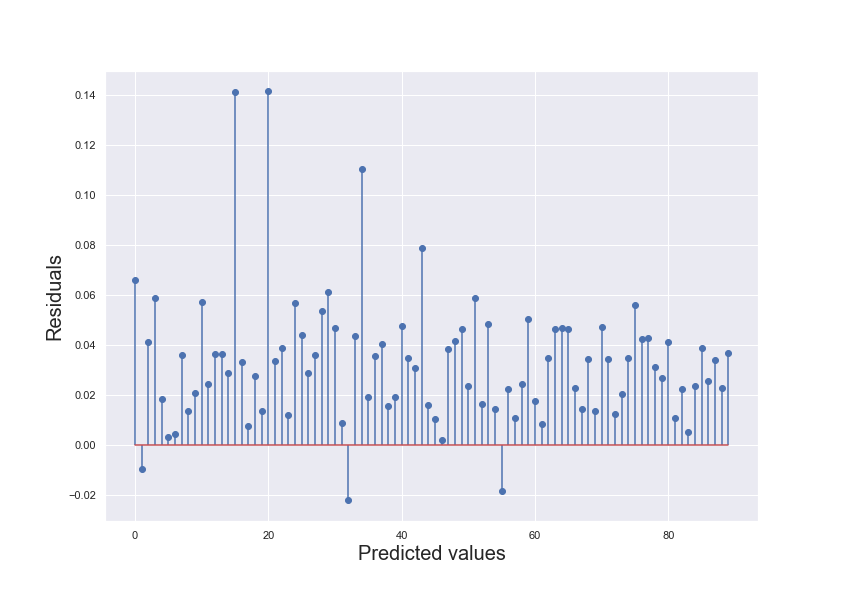}
  \caption{}
  \label{fig:sub1}
\end{subfigure}%

\begin{subfigure}{0.9\textwidth}
  \centering
  \includegraphics[width=0.95\linewidth]{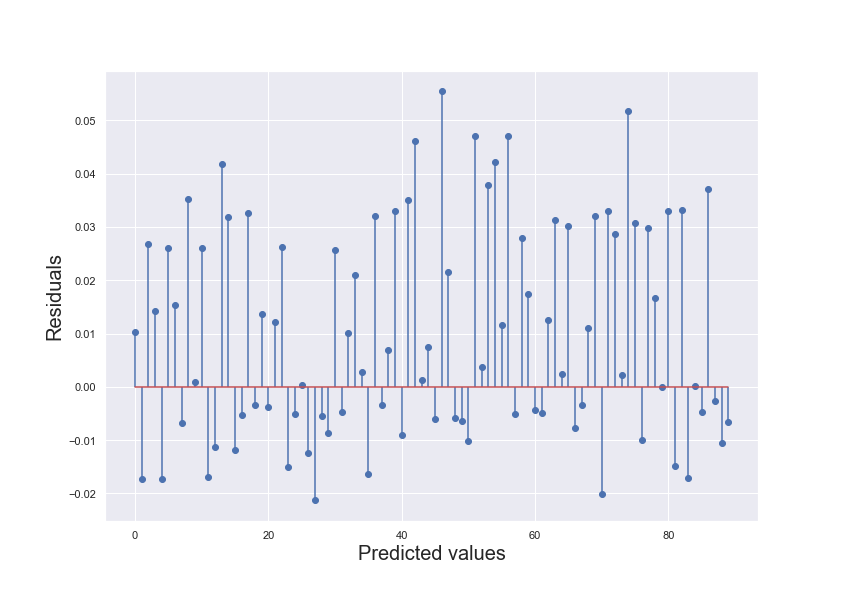}
  \caption{}
  \label{fig:sub2}
\end{subfigure}
\caption{K=2, distanța euclidiană}
\label{fig:test}
\end{figure}

\begin{figure}
\hspace*{-2.5cm}
\centering
\begin{subfigure}{0.65\textwidth}
  \centering
  \includegraphics[width=1\linewidth]{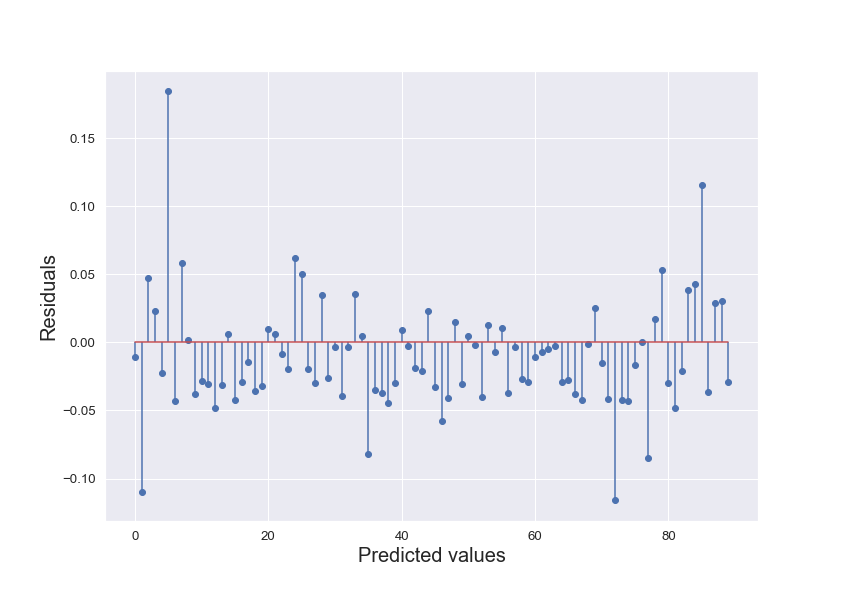}
  \caption{}
  \label{fig:sub1}
\end{subfigure}%
\begin{subfigure}{0.65\textwidth}
  \centering
  \includegraphics[width=1\linewidth]{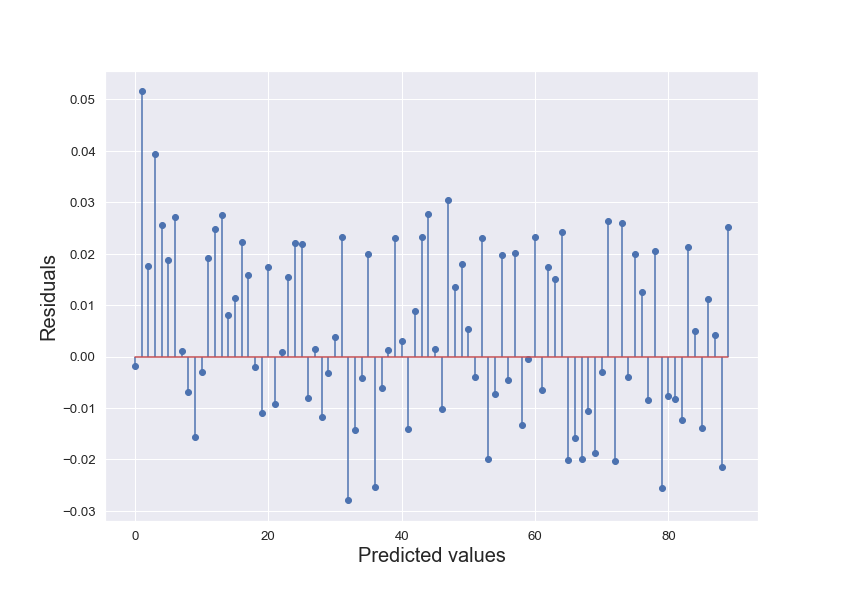}
  \caption{}
  \label{fig:sub2}
\end{subfigure}

\hspace*{-2.5cm}
\begin{subfigure}{0.65\textwidth}
  \centering
  \includegraphics[width=1\linewidth]{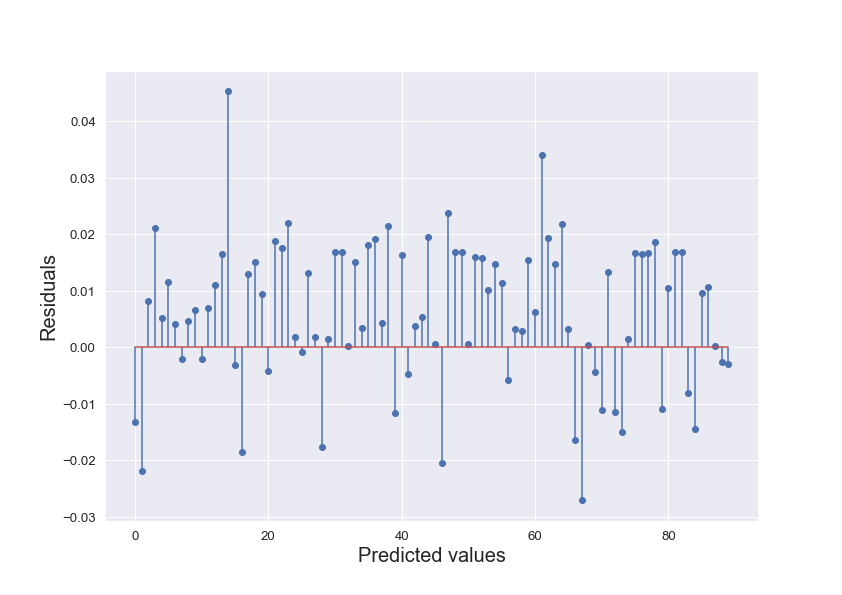}
  \caption{}
  \label{fig:sub1}
\end{subfigure}%
\begin{subfigure}{0.65\textwidth}
  \centering
  \includegraphics[width=1\linewidth]{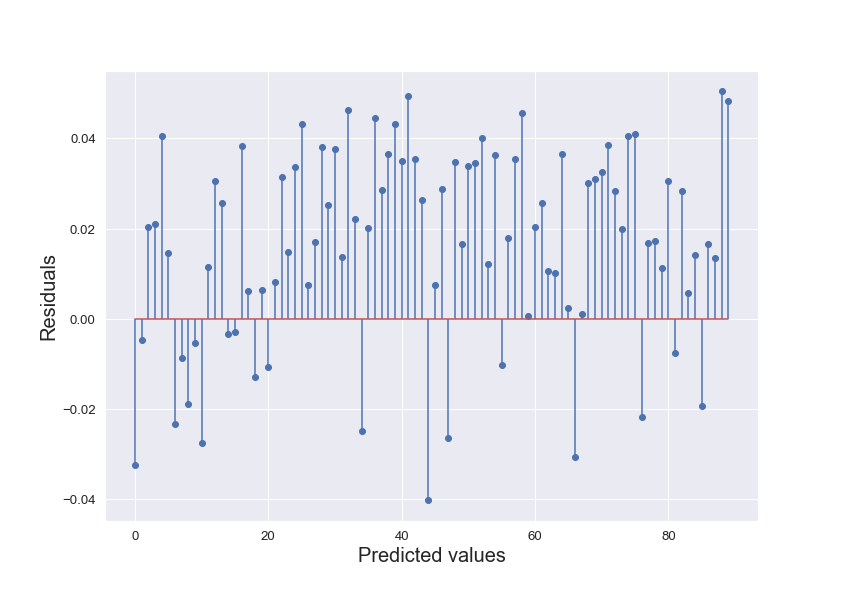}
  \caption{}
  \label{fig:sub2}
\end{subfigure}
\caption{K=4, distanța euclidiană}
\label{fig:test}
\end{figure}

\begin{figure}
\centering
\begin{subfigure}{0.9\textwidth}
  \centering
  \includegraphics[width=0.95\linewidth]{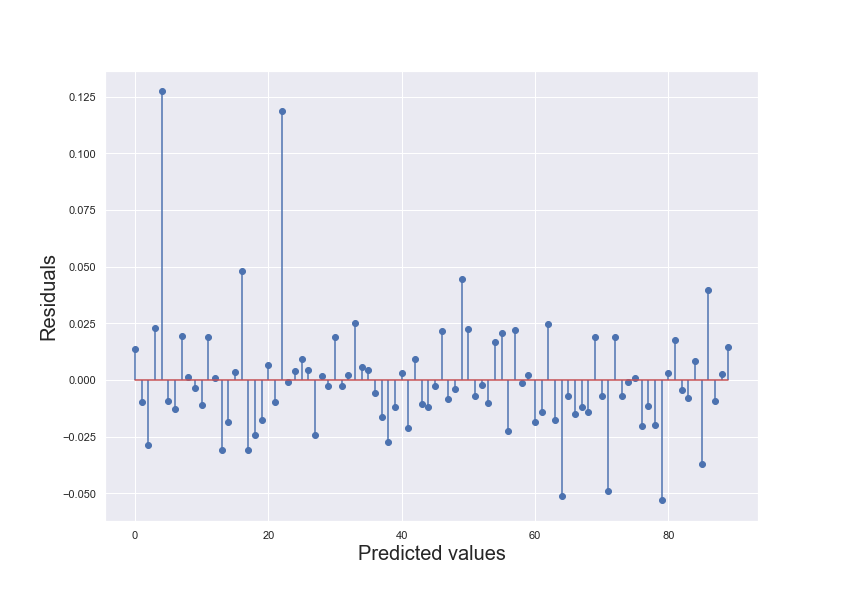}
  \caption{}
  \label{fig:sub1}
\end{subfigure}%

\begin{subfigure}{0.9\textwidth}
  \centering
  \includegraphics[width=0.95\linewidth]{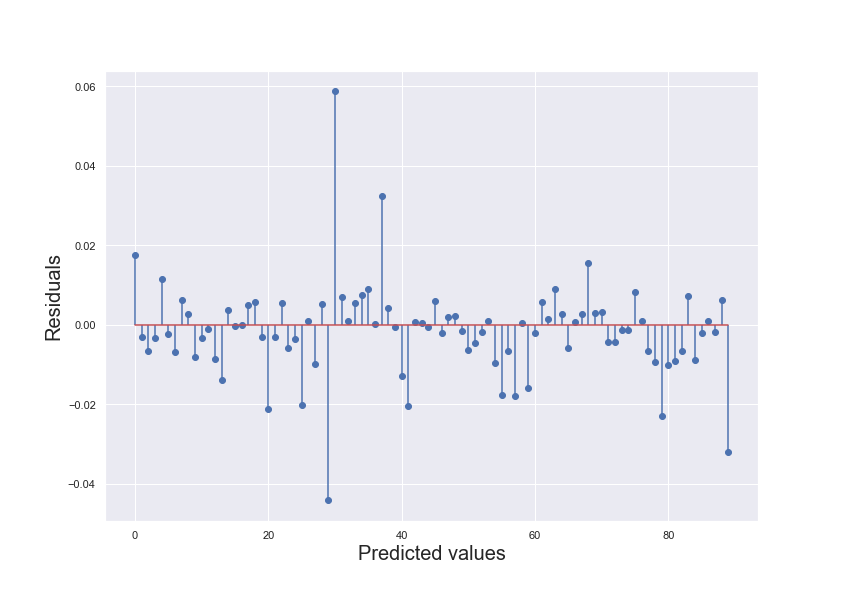}
  \caption{}
  \label{fig:sub2}
\end{subfigure}
\caption{K=2, distanța manhattan}
\label{fig:test}
\end{figure}

\begin{figure}
\hspace*{-2.5cm}
\centering
\begin{subfigure}{0.65\textwidth}
  \centering
  \includegraphics[width=1\linewidth]{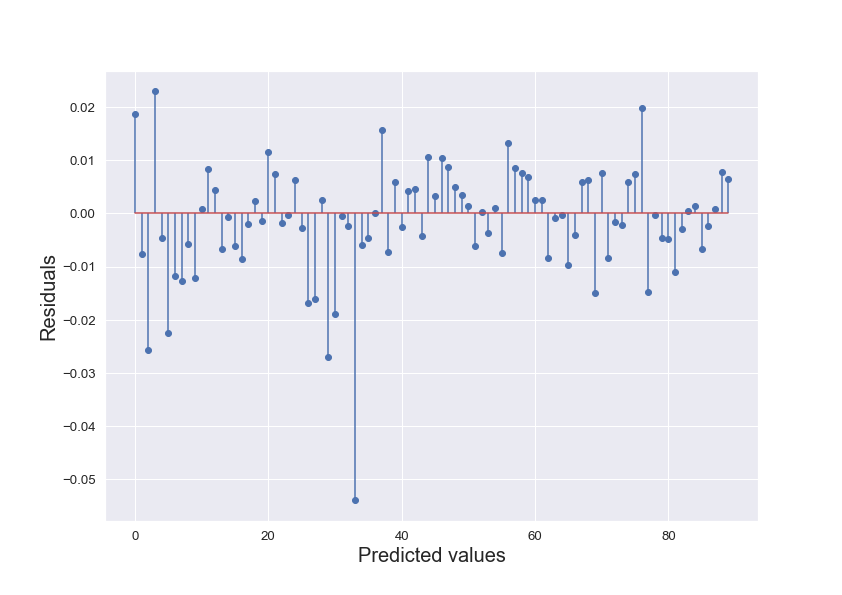}
  \caption{}
  \label{fig:sub1}
\end{subfigure}%
\begin{subfigure}{0.65\textwidth}
  \centering
  \includegraphics[width=1\linewidth]{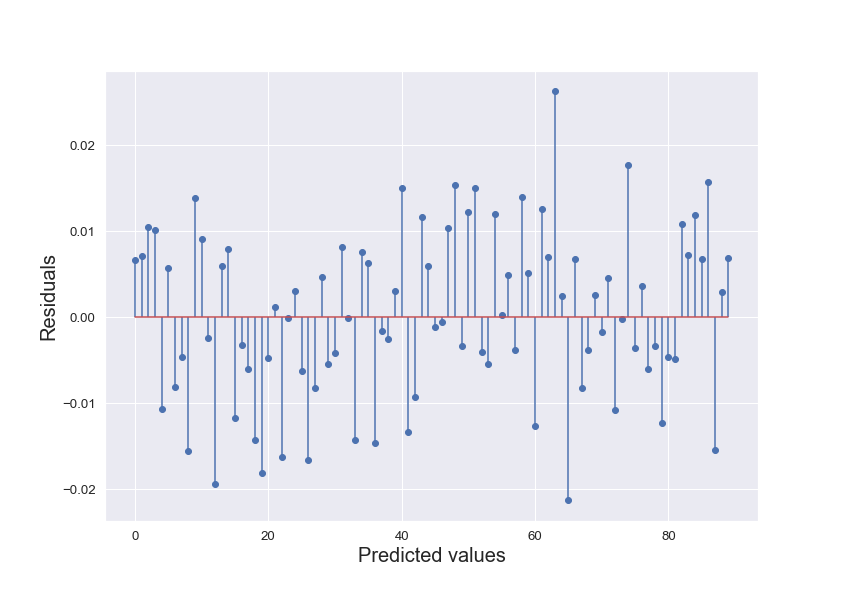}
  \caption{}
  \label{fig:sub2}
\end{subfigure}

\hspace*{-2.5cm}
\begin{subfigure}{0.65\textwidth}
  \centering
  \includegraphics[width=1\linewidth]{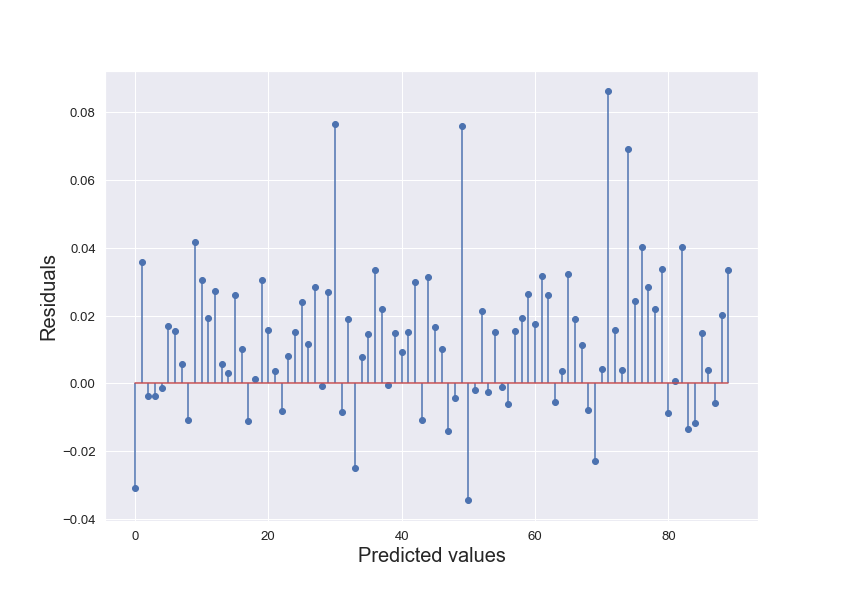}
  \caption{}
  \label{fig:sub1}
\end{subfigure}%
\begin{subfigure}{0.65\textwidth}
  \centering
  \includegraphics[width=1\linewidth]{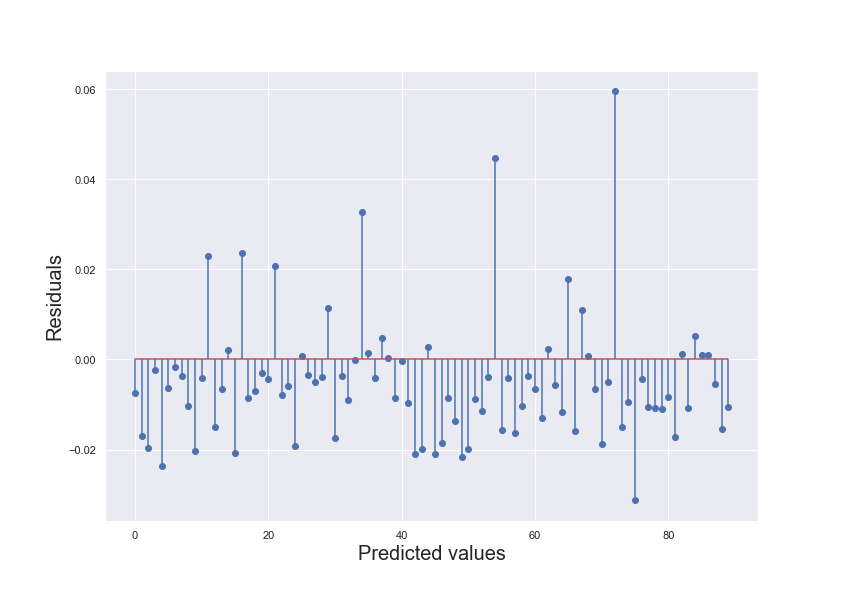}
  \caption{}
  \label{fig:sub2}
\end{subfigure}
\caption{K=4, distanța manhattan}
\label{fig:test}
\end{figure}
\newpage
\newpage
\newpage
\newpage

Pentru predicatul bazat pe distanța Minkowski, am ales valori pentru $p=1$ și $p=2$ pentru a vedea dacă rezultatele sunt asemănătoare cu cele obținute pentru distanța Manhattan și Euclidiană.

\begin{table}[!htb]
    \caption[K=2, distanța minkowski]
    {K=2, distanță minkowski\\pentru (a),(b) $p=1$\\pentru (c),(d) $p=2$}
    \begin{subtable}{0.1\linewidth}
      \centering
        \caption{}
        \begin{tabular}{|l|l|l|l|l|l|}
\hline
y       & 0.98  & 0.98  & 0.12 & 0.466 & 0.073 \\ \hline
y\_pred & 0.925 & 0.94 & 0.157 & 0.502 & 0.107 \\ \hline
dif     & 0.055 & 0.04 & 0.037 & 0.036 & 0.034 \\ \hline
\end{tabular}
    \end{subtable}%
    \begin{subtable}{1.45\linewidth}
      \centering
        \caption{}
\begin{tabular}{|l|l|l|l|l|l|}
\hline
y       & 0.98  & 0.687  & 0.466 & 0.76 & 0.69 \\ \hline
y\_pred & 1.021 & 0.722 & 0.499 & 0.792 & 0.722 \\ \hline
dif     & 0.041 & 0.035 & 0.033 & 0.032 & 0.032 \\ \hline
\end{tabular}
    \end{subtable} 
        \begin{subtable}{0.1\linewidth}
      \centering
        \caption{}
        \begin{tabular}{|l|l|l|l|l|l|}
\hline
y       & 0.414  & 0.563  & 0.56 & 0.59 & 0.29 \\ \hline
y\_pred & 0.376 & 0.532 & 0.533 & 0.567 & 0.269 \\ \hline
dif     & 0.038 & 0.031 & 0.027 & 0.023 & 0.021 \\ \hline
\end{tabular}
    \end{subtable}%
    \begin{subtable}{1.45\linewidth}
      \centering
        \caption{}
\begin{tabular}{|l|l|l|l|l|l|}
\hline
y       & 0.98  & 0.98  & 0.29 & 0.59 & 0.946 \\ \hline
y\_pred & 1.027 & 1.01 & 0.261 & 0.562 & 0.972 \\ \hline
dif     & 0.047 & 0.03 & 0.029 & 0.028 & 0.026 \\ \hline
\end{tabular}
    \end{subtable} 
\end{table}

Pentru cele doua valori, am obținut rezultate chiar mai bune decât pentru predicatele anterioare. Pentru valoarea \textbf{0.98}, aleasă deoarece rezultatele în prezicerea acestei poziții au fost cele mai slabe, diferența absolută dintre poziția prezisă și cea reală sunt următoarele:

\begin{itemize}
    \item predicatul manhattah - 0.127
    \item predicatul euclidian - 0.142
    \item predicatul minkowski(p=1) - 0.055
    \item predicatul minkowski(p=2) - 0.047
\end{itemize}

Rezultatele acestui experiment demonstrează că o rețea LTN poate fi folosită cu succes chiar și pe date reale pentru a prezice poziția deteriorării pe o grindă de metal, cu beneficiul ușurinței schimbării predicatului de distanță/similaritate. Prin compararea rezultatelor cu cele din \cite{gillich2022beam}, LTN oferă o precizie mult mai bună decat algoritmii de învățare automată implementați în lucrarea respectivă.



\chapter{Interpretarea rezultatelor și concluzionarea lucrării}

\section{Avantaje/Dezavantaje ale celor două abordări}
În această parte a lucrării sunt discutate avantajele și dezavantajele rețelei neuro-symbolice și rețelei neuronale profunde, din urma realizării experimentelor anterioare.

\textbf{Rețeaua neuronală profundă}\\
\textbf{Avantajele unei rețele neuronale profunde}
\begin{itemize}
    \item Este nevoie de puține schimbări, în unele cazuri chiar nefiind necesare, pentru a o folosi pentru diferite aplicații sau tipuri de date.
    \item Numeroase articole din surse de încredere arată rezultate foarte bune în predicții, recunoașteri și clasificări.
    \item În funcție de modul în care este folosită rețeaua, ea poate să rezolve probleme supervizată total, fără supervizare, supervizată parțial sau prin armare. Aceste moduri acoperă majoritatea provocărilor din Inteligența Artificială.
\end{itemize}

\textbf{Dezavantajele unei rețele neuronale profunde}
\begin{itemize}
    \item Lipsa transparenței asupra rezultatelor modelului. Este greu de înțeles cum modelul a produs din datele de intrare, datele de ieșire. Problema "Black Box" reprezintă unul dintre principalele motive din spatele interesului crescut asupra sistemelor hibride.
    \item Pentru a implementa chiar și o rețea neuronale profundă simplă, este nevoie de o bună înțelegere asupra elementelor de statistică, probabilități, algebră liniară, calcul diferențial, calcul integral și analiza datelor.
    \item Este o șansă mare ca modelul să nu poată generaliza. Antrenarea să se concentreze pe detalii irelevante și să nu facă conexiunile necesare între datele oferite. Rezultatul este o performață slabă pe date noi.
\end{itemize}

\textbf{Rețeaua neuro-symbolică}\\
\textbf{Avantajele Logic Tensor Networks}
\begin{itemize}
    \item Logic Tensor Networks poate să rezolve majoritatea problemelor din Inteligența Artificială. Implementări ale problemelor de clasificare binară, clasificare single-label, clasificare multi-label, clustering și predicție sunt oferite de autorii rețelei neuro-symbolice LTN \cite{badreddine2022logic}. O parte din acestea fiind folosite și în realizarea experimentelor acestei lucrări.
    \item Sistemele hibride au nevoie de un volum mai mic de date pentru antrenare, datorită faptului că logica este definită încă de la îceput urmând să fie folosită împreuna cu logica dobândită din antrenare. Acest lucru face ca rețeaua neuro-symbolică să aibă rezultate mai bune chiar din primele epoci ale antrenării. Relațiile dintre obiecte, inferențele logice și cunoașterea oncologică pot să fi definite prin Real Logic și pot face antrenarea mai eficientă.
    \item Probabil cel mai important aspect al LTN este partea de raționament logic care face posibil pentru oameni să înțeleagă cum un model produce un anumit rezultat. Se pot face interogări ale nivelului de satisfiabilitate al unei formule, interogări ale constrângerilor și se poate măsura acuratețea. Acuratețea modelului este influențată direct de nivelul de satisfiablitate al bazei de cunoștințe, iar o predicție incorectă duce la o scăderea satisfiablității și modificarea parametrilor componentelor definite prin Real Logic.
    
    \item Problema de supraadaptarea(overfitting) este mitigată prin adăugarea constrângerilor în baza de cunoștințe.

\end{itemize}

\textbf{Dezavantaje Logic Tensor Networks}
\begin{itemize}
    \item Cu toate ca LTN poate să rezolve probleme precum o clasificare multiclass, această abordare poate să nu fie cea mai bună. Drept exemplu, setul de date KDD99 folosit în această lucrare are 42 de tipuri de atacuri care, în cazul ideal, ar trebui adaugate în baza de cunoștințe pentru a putea fi clasificate adecvat. Cu siguranță, acest lucru este posibil, precum și adăugarea relațiilor dintre atributele fiecărei conexiuni și fiecare atac în parte, dar implementarea nu este una ușoară.  
    \item Unele probleme pot avea nevoie de set mare de axiome. De remarcat este faptul că volumul setului de axiome afectează puterea de calcul necesară.
\end{itemize}

Tabela 5.1 de mai jos prezintă, într-o formă compactă, avantajele și dezavantajele celor două abordări. 

\newpage

\begin{table}[!htbp]
\begin{tabular}{|ll|}
\hline
\multicolumn{2}{|c|}{\textbf{Rețea neuronală profundă}}                               \\ \hline
\multicolumn{1}{|l|}{\textbf{PROS}}               & \textbf{CONS}                \\ \hline
\multicolumn{1}{|l|}{\begin{tabular}[c]{@{}l@{}}implementări flexibile pentru \\ probleme viitoare\end{tabular}} &
  black box \\ \hline
\multicolumn{1}{|l|}{rezultate foarte bune demonstrate} &
  \begin{tabular}[c]{@{}l@{}}necesită o întelegere aprofundata\\ a învățării automate \end{tabular} \\ \hline
\multicolumn{1}{|l|}{poate rezolva majoritatea problemelor din AI} & supraantrenarea poate apărea \\ \hline
\end{tabular}
\end{table}
\vspace{-0.65cm}
\begin{table}[!htbp]
\begin{tabular}{|ll|}
\hline
\multicolumn{2}{|c|}{\textbf{Logic Tensor Network}} \\ \hline
\multicolumn{1}{|l|}{\textbf{PROS}} &
  \textbf{CONS} \\ \hline
\multicolumn{1}{|l|}{rezolvă majoritatea problemelor din AI} &
  \begin{tabular}[c]{@{}l@{}}unele probleme necesită\\  o bază mare de cunoștințe\end{tabular} \\ \hline
\multicolumn{1}{|l|}{necesită volum mai mic al datelor de antrenare} & \begin{tabular}[c]{@{}l@{}}probleme de scalabilitate prin adăugarea\\ mai multor formule în setul de axiome\end{tabular} \\ \hline
\multicolumn{1}{|l|}{\begin{tabular}[c]{@{}l@{}}acuratețe interactivă, învățarea \\ și deducția logică devin posibile \end{tabular}} & -
   \\ \hline
\multicolumn{1}{|l|}{\begin{tabular}[c]{@{}l@{}}supraantrenarea poate fi evitată cu ușurință \end{tabular}} & -
   \\ \hline
\end{tabular}\\
\caption{\label{tab:table-name}}
\end{table}
\vspace{-0.5cm}

\section{Dezvoltări ulterioare}
Cu toate că în această lucrare rezultatele obținute sunt foarte bune, bazele de cunoștințe și constrângerile realizate și interogate sunt simple. O posibilă lucrare viitoare ar putea adresa acest aspect și ar putea să încerce limitele logicii de ordin întâi definită prin Real Logic.

La momentul scrierii acestei lucrări, trei viitoare articole pe teme similare urmează a fi publicate:
\begin{itemize}
    \item Advantages of a neuro-symbolic solution for
monitoring IT infrastructures alerts - \textbf{SYNASC 2022}
    \item A neuro-symbolic solution for observing network traffic
alerts - \textbf{Expert Systems With Applications, Elsevier}
    \item A neuro-symbolic model for cantilever beams damage detection - \textbf{Computers in Industry, Elsevier}
\end{itemize}

\section{Concluzie}

Rețeaua hibridă bazată pe Logic Tensor Networks s-a dovedit, în urma experimentelor realizate, superioară în recunoașterea atacurilor din seturile de date KDD99 și CIC-IDS2017, printr-o comparație directă a acestor rezultate și cele obținute prin soluții bine știute precum rețele neuronale profunde. În cazul ultimului experiment, rezultatele au fost comparate cu cele din lucrarea din care provine și setul de date folosit și a fost demonstrată o diferență mare în precizie. Toate acestea, precum și discuția despre avantajele și dezavantajele celor două abordări, justifică adăugarea unei părți de raționament logic asupra algoritmilor de învățare profundă și încurajează dezvoltările acestei noi ramuri a inteligenței artificiale.

\printbibliography

\end{document}